\documentclass[manuscript,screen]{acmart}

\usepackage{verbatim}
\usepackage{graphicx}
\usepackage{url} 
\usepackage{listings}
\usepackage{pslatex} 
\usepackage{hyperref}
\usepackage{amsmath}
\usepackage{CJKutf8}
\usepackage{booktabs}
\usepackage{subcaption}
\usepackage{pdfpages}
\AtBeginDocument{%
  }

\setcopyright{acmlicensed}
\copyrightyear{2018}
\acmYear{2018}
\acmDOI{XXXXXXX.XXXXXXX}

\acmConference[Conference acronym 'XX]{Make sure to enter the correct
  conference title from your rights confirmation email}{June 03--05,
  2018}{Woodstock, NY}
\acmISBN{978-1-4503-XXXX-X/18/06}

\begin{document}

\title{\textsc{CantonMT}: Investigating Back-Translation and Model-Switch Mechanisms for Cantonese-English Neural Machine Translation}

\author{Kung Yin Hong}
\email{kenrick.kung@gmail.com}
\affiliation{%
  \institution{University of Manchester}
  \city{Greater Manchester}
  \state{England}
  \country{UK}
}

\author{Lifeng Han}
\authornote{Corresponding author.}
\affiliation{%
  \institution{University of Manchester}
  \city{Greater Manchester}
  \state{England}
  \country{UK}
}
\email{Lifeng.Han@manchester.ac.uk}

\author{Riza Batista-Navarro}
\affiliation{%
  \institution{University of Manchester}
  \city{Greater Manchester}
  \state{England}
  \country{UK}
}
\email{riza.batista.Han@manchester.ac.uk}

\author{Goran Nenadic}
\affiliation{%
  \institution{University of Manchester}
  \city{Greater Manchester}
  \state{England}
  \country{UK}
}
\email{g.nenadic@manchester.ac.uk}

\renewcommand{\shortauthors}{Hong et al.}
\renewcommand{\shorttitle}{\textit{\textsc{CantonMT}: Investigating Back-Translation and Model-Switch Mechanisms for Cantonese-English NMT}}

\begin{abstract}
This paper investigates the development and evaluation of machine translation models from Cantonese to English, where we propose a novel approach to tackle low-resource language translations.
Despite recent improvements in Neural Machine Translation (NMT) models with Transformer-based architectures, Cantonese, a language with over 80 million native speakers, has below-par State-of-the-art commercial translation models due to a lack of resources. The main objectives of the study are to develop a model that can effectively translate Cantonese to English and evaluate it against state-of-the-art commercial models.
To achieve this, a new parallel corpus has been created by combining different available corpora online with preprocessing and cleaning. In addition, a monolingual Cantonese dataset has been created through web scraping to aid the synthetic parallel corpus generation. 
Following the data collection process, several approaches, including fine-tuning models, back-translation, and model switch, have been used. The translation quality of models has been evaluated with multiple quality metrics, including lexicon-based metrics (SacreBLEU and hLEPOR) and embedding-space metrics (COMET and BERTscore). Based on the automatic metrics, the best model is selected and compared against the 2 best commercial translators using the human evaluation framework HOPES.
The best model proposed in this investigation (NLLB-mBART) with model switch mechanisms has reached comparable and even better automatic evaluation scores against State-of-the-art commercial models (Bing and Baidu Translators), with a SacreBLEU score of 16.8 on our test set. 
Furthermore, an open-source web application has been developed to allow users to translate between Cantonese and English, with the different trained models available for effective comparisons between models from this investigation and users. \textsc{CantonMT} is available at \url{https://github.com/kenrickkung/CantoneseTranslation}
\end{abstract}



\received{xx May 2024}
\received[revised]{xx 2024}
\received[accepted]{xxx}

\maketitle

\section{Introduction}
Cantonese is a Sinitic language spoken in Hong Kong, Macau, and the Guangdong region of southern PRC, it is the second most spoken Sinitic language, after Mandarin Chinese  \citep{wiedenhof2015grammar}. With a substantial 80 million native speakers  \citep{eberhard-2021-ethnologue}, Cantonese is still an under-researched area in the spectrum of Natural Language Processing, as demonstrated in ACL Anthology, where only 47 papers are related to Cantonese, compared with 2355 for (Mandarin) Chinese  \citep{xiang2022cantonese}.

Despite having the second most speakers in the family of Sinitic languages, most State-of-the-art commercial translators either do not support Cantonese or have below-par translation quality when translated to English. This leads to scenarios where individuals seeking Cantonese resources face challenges, particularly in casual forums where tones are often very similar to spoken language. 

We believe that Cantonese is a unique language that captures the rich cultural history of Hong Kong, Macau, and the Guangdong province of China. Two major challenges when dealing with Cantonese translations are Colloquialism and Multilingualism. Colloquialism, a linguistic style used for informal and casual conversation, often occurs in Cantonese and includes non-standard spelling, slang, and neologisms. 
As for Multilingualism, Hong Kong was once a British colony and has a rich Chinese cultural influence; code-switching \footnote{the act of using multiple languages together} happens often in day-to-day conversation; and words can also be loaned from English through phonetic transliteration  \citep{robert2006loan}. 
Therefore, following the trend of language diversity and inclusion in NLP, we have set out the aim to develop a translation system that could translate sentences from Cantonese to English and reach comparable results against commercial translators.

In the era of the fast development of  {NLP}, many  {MT} models have been proposed for the majority of languages worldwide. However, low-resource language  {MT} still challenges researchers. Cantonese can certainly be viewed as one of the low-resource languages given that its written form of collection is scarce. It is well known that  {NMT} models require much data to obtain good translation quality. In this work, we aim to investigate one of the popular methods, i.e. synthetic data generation for data-augmented fine-tuning via forward-translation and back-translation models on Cantonese-to-English  {NMT}.

This work aims to fill the gap in the NLP and MT field for an open-sourced Cantonese-to-English translation tool that can obtain comparable results against commercial translators, which could potentially allow users to use the system locally. 
With this in mind, we set the following objectives:

\begin{itemize}
  \item Create Parallel and Monolingual corpus that can be used in further research on Cantonese  {NLP}
  \item Develop state-of-the-art models for translating Cantonese to English 
  \item Create an open-sourced User Interface that can be used for translation and as a toolkit for future projects on translation task
\end{itemize}

Regarding the Evaluation Strategy, the models developed are evaluated through a range of metrics, including lexicon-based word surface matchings (SacreBLEU and hLEPOR) and those based on embedding spaces (COMET and BERTscore). Following these metrics, the top-performing model is chosen for comparison with the two top-performing commercial translation tools, employing the HOPES human evaluation framework, which we modified based on HOPE, a human-centric post-editing based metric  \cite{gladkoff-han-2022-hope}. \footnote{This is an extended solid investigation based on our preliminary work reported in  \cite{hong2024cantonmt}. }
The rest of the paper is organised as below:
\begin{itemize}
    \item Section \ref{cha:tech-bg} introduces some technical backgrounds and related works on MT, LLMs, back-translation, data augmentation, and Cantonese NLP. 
    \item Section \ref{cha:Method} presents the methodology and model design for \textsc{CantonMT} including corpus collection, data preprocessing, model selections, training and fine-tuning, plug-and-play / model-switch mechanism. 
    \item Section \ref{cha:experiments} lays out the experimental evaluation metrics, human evaluations, and analyses of model outcomes and their comparisons to state-of-the-art commercial translators. 
    \item Section \ref{cha:conclude} concludes this paper with findings and future work plans.
\end{itemize}

\section{Background and Related Works}
\label{cha:tech-bg}
\subsection{Traditional Machine Translation}
 {MT} has long been one of the most important tasks in  {NLP} intending to translate sentences into another language with the help of a computer. It was first introduced by  \citet{weaver1952translation}.
 Rule-based MT (RBMT), also known as Knowledged-Based MT, is one of the classical approaches to MT. It was first developed in the early 1970s, and one of the main systems was SYSTRAN \footnote{\url{http://www.systran.de/}}. Having an input sentence (in the source language), an  {RBMT} system generates them to output sentence (in the target language) based on morphological, syntactical, and semantic analysis on both the source and the target language. 
This approach had considerable merit at the time; however, it has its shortcomings. For example, building a dictionary for a new language is very expensive, and rules are handwritten, which could mean a lot of human judgement might be involved. Therefore, it has been replaced by Corpus-based MT and Statistical MT (SMT)  \citep{brown1988SMT,koehn2003statistical}. 
SMT is based on the availability of pairs of corresponding text that are translations of each other (Parallel Corpus) and does not rely on a deep understanding of the grammatical rules of the languages involved, which means the machine can automatically analyse the parallel corpus and generate translated text without the need of hand-crafted rules, via bilingual ``phrase table'' and statistical ``language model''.

However,  {SMT} struggles with a few issues, where proper nouns might be overridden due to statistical anomaly. An example would be a sentence such as "I took a flight to Berlin" might be translated to "I took a flight to Paris" instead due to the abundance of "to Paris" in the training data. Some other issues that it may struggle with are word order, idioms, and the inability to model long-distance dependencies between words  \citep{han2022investigation}.

\subsection{Neural Machine Translation}
With the breakthrough in computing power and deep learning, neural networks have become the paradigm to  {MT}, which will be  investigated in this project.
These models are defined as  {NMT} and it is a radical change from the previous  {MT} approaches where it does not require additional feature engineering and is instead an end-to-end training process, which simplifies the need for language-dependent expertise. 

 {Seq2Seq} Models  \citep{suts2014seq2seq,cho2014seq} are a specific realisation of  {RNN} models and are particularly useful for  {MT} since they take a sequence (source sentence) as input and output a sequence (target sentence). It is usually composed of an encoder and a decoder, where the encoder captures the context and meaning of the input sequence, and the decoder uses the encoded input to produce a final output sequence. In the early paper when this architecture was first introduced, the encoder and decoder are normally two {RNN}s.

However, despite the advancements made, this architecture struggles with information-management bottleneck, which may lead to information loss over long-distance language dependencies. 
The issues rise as for the encoder, it is difficult to encode the meaning of the sentence in a vector representation, and for the decoder, some information might be more relevant than others at different parts of the output sequence.

The attention mechanism was introduced into  {RNN} by  \citet{Bad2014Attention}, where it aims to solve the information bottleneck from vanilla  {RNN}s. 
The Transformer architecture was later introduced based on the attention mechanism  \citep{vaswani2017transformer} and has outperformed the RNN-based models. 

\subsection{Large Language Models}
 {LLMs} were first introduced by  \citet{devlin2019bert} with the introduction of  {BERT} and can be use for general purpose  {NLP} tasks. It usually will be trained in two phases, pre-training and fine-tuning. Since the introduction of  {LLM}s, they have swept the leaderboard for most  {NLP} tasks and has reached new state-of-the-art results for probably all main  {NLP} tasks including  {MT}.

With the rise of  {LLM}s, there are dozens of pre-trained models which are capable on  {MT} tasks with none or few fine-tuning. In our investigation, there are 3 models chosen for further fine-tuning with our dataset, the reason behind choosing these models can be seen in the methodology section. Here is a brief introduction of each model, which could help readers understand the difference with depth.
\subsubsection{Opus-MT}
Opus-MT  \citep{Tiedemann2020Opus}, developed by Helsinki-NLP, is a Transformer-based  {NMT}, which is using Marian-NMT \footnote{\url{https://marian-nmt.github.io/}} as the framework for the model training. The model family is trained with a publicly available parallel corpus collected in OPUS\footnote{\url{http://opus.nlpl.eu/}}. The model is specifically trained for  {MT} task, and should not be classified as a general purpose {LLM}.

Two specific models are used in this project, \textit{Opus-mt-zh-en} and \textit{Opus-mt-en-zh}, which are models that translate Chinese to English and English to Chinese. The forward model (Chinese to English) has around 77M parameters, which is considered quite a small model when compared to {LLM}s.

\subsubsection{mBART}
mBART  \citep{liu2020mbart}, a multilingual  {Seq2Seq} denoising auto-encoder. It is trained with the BART  \citep{lewis2020bart} objectives with a multilingual corpus.

The pre-training of mBART is trained by corrupting text with a noising function and also learning a model to reconstruct the original text. It uses the CC25 Corpus which contains 25 languages and follows the standard Transformer architecture with 12 layers of encoders and 12 layers of decoders.

In this paper, a specific version of the model is used (\textit{mbart-large-50-many-to-many-mmt}) which supports 50 languages, including (Mandarin) Chinese. However, it does not support Cantonese as a language. The model is also fine-tuned for multilingual translation and is introduced by  \citet{tang2020multilingual} which has added 25 additional languages without hurting the performance of the model. The model has a total of 610M parameters, a massive increase compared to the previous Opus model.

\subsubsection{NLLB}
No Language Left Behind (NLLB)  \citep{nllbteam2022language}, to the best of our knowledge, is the only publicly available  {LLM} which contains the language Cantonese (Lang-Code: yue\_Hant). It is trained upon the FLORES-200 dataset which contains 200 languages and serves as a high-quality benchmark dataset. The model architecture is also based on the Transformer encoder-decoder architecture  \citep{vaswani2017transformer}.

In this work, a distilled version of NLLB (\textit{nllb-200-distilled-600M}) is used since based on our available computation power, there is no chance of fine-tuning a larger model. The model is already fine-tuned on  {MT} task, and the language pair in focus is Cantonese-English.

\subsection{Back-Translation}
Data Augmentation via Back translation is a technique used by  {MT} researchers when tackling low-resource languages. Typically, since not enough data is available, the model may not be able to learn the translation of the language thoroughly and, thus might harm the performance of  {MT}.
This technique has been one of the standards for leveraging monolingual corpora since  {SMT} \citep{bojar2011backtranslationsmt}, and is still being used with {NMT}  \citep{sennrich2016backtransnmt}. 

The approach uses a model, which translates target language text to the source language (back model), for translating a monolingual corpus in the target language to the source language. This creates a synthetic parallel corpus (Silver Standard), which is different from human annotated parallel corpus (Gold Standard). In theory, with more data, the model can be performing better.

Iterative approaches are also being explored in this project  \citep{hoang2018iterative}. 
A more in-depth explanation of the approach taken in this project is given in the methodology section.
Below are some works of literature that follow this technique. 

\subsubsection{English-Vietnamese Translation}
For back-translation, a monolingual corpus is required, often involving web scraping, which may lead to low-quality datasets since text from the internet often does not follow proper grammar rules. In light of that,  \citet{pham2023engviet} has proposed a method to correct the sentence with grammatical errors to improve the quality of monolingual corpus.
\subsubsection{Context-aware Neural Machine Translation}
 \citet{sugiyama2019contextBT} developed a context-aware {NMT} model which translates Japanese to English via back translation. They have reached state-of-the-art results in single-sentence translation from Japanese to English and Japanese to French.
\subsubsection{English-German Translation}
 \citet{graca2019engger}  pointed out that sampling-based synthetic data generation schemes have some fundamental problems and proposed methods to remedy them by disabling label-smoothing for the back model and sampling from a restricted search space.
In our project, due to the nature of Cantonese translation, the methodology proposed by this work is not likely to succeed, and the approach in the project is a different approach inspired by  \citet{hoang2018iterative}, which will be explained in the methodology section.

\subsection{MT on Cantonese}
\subsubsection{Commercial Translators}
A survey has been conducted on four different commercial  {MT} software, including Google, Bing, Baidu, and DeepL.

For Google\footnote{\url{https://translate.google.com/}} and DeepL\footnote{\url{https://www.deepl.com/translator}}, despite being the most popular software used for translation in daily lives, they do not support Cantonese as an option, but only (Mandarin) Chinese. Therefore, no further investigations are being made on the platforms.
For Bing\footnote{\url{https://www.bing.com/translator}} and Baidu\footnote{\url{https://fanyi.baidu.com}}, there are native Cantonese support in translation and therefore are chosen as a state-of-the-art comparison in the following sections.

With the rise of  {LLM}s, there are also questions on whether or not this kind of model with prompting can give better results when compared with a more traditional approach with fine-tuning on  {LLM}s. In this project, Generative Pre-trained Transformers(GPT)-4  \citep{openai2024gpt4} are being investigated with specific prompting to compare against our model. 
The implementation of GPT-4 that we used is Cantonese Companion, which was custom-made for translation to Cantonese by a community builder.\footnote{\url{https://chat.openai.com/share/7ee588af-dc48-4406-95f4-0471e1fb70a8}}
However, it should be noted that we do not know how much data was used for this community-trained Cantonese Companion and the training was not transparent, in addition to its dependence on the commercial platform.

\subsubsection{Research Models}
Research work focusing on Cantonese-English  {MT} has not gained much attention up to date unfortunately, and therefore to expand the search,  {MT} models that are focusing on Cantonese have all been investigated to understand the frontier of the research in Cantonese translation.

\textbf{Example-based and Rule-based Machine Translation}
 \citet{wu2006structural} has presented a method for Cantonese-English  {MT} with a combination of example-based and  {RBMT}, which uses morphological knowledge, syntax analysis, translation examples and target-generation-based rules. Their approach has reached 80\% accuracy on their test data report. However, this approach requires advanced knowledge in both Cantonese and English, which is very hard to scale up since it requires a lot of feature engineering and also domain-specific knowledge.

\textbf{Neural Machine Translation}
 \citet{wing2020machine} has created a plan to develop different models in Cantonese-English  {MT}, which include  {RNN} and also Transformers model, however, the result cannot be found anywhere online, and therefore no cross-comparison can be used.

To the best of our knowledge, these are the only literature which has attempted to tackle the translation direction from Cantonese to English. There is also one piece of literature in the opposite direction, which we have taken inspiration from the project in terms of the dataset used which we will explain next.

\textbf{English-Cantonese Machine Translation}
TransCan\footnote{\url{https://github.com/ayaka14732/TransCan}} is a  {NMT} model which translates English to Cantonese and is trained based on bart-base-Chinese and BART with additional linear projection to connect them; it has reached 28.6 in BLEU score on the tested data which is much higher than the available commercial translation. However, due to the nature of the BLEU score and also it being a different direction, the result is not comparable to the model we will be proposing.

There is also literature on \textit{Chinese-Cantonese  {NMT}} in a looser direction. They often use the word Mandarin instead of Chinese since Chinese could also be an umbrella term covering both Mandarin and Cantonese. However, to distinguish the difference between Chinese and Cantonese here, the report will use Chinese interchangeably with Mandarin. 

\textbf{BiLSTM Machine Translation}
 \citet{liu-2022-low} has developed a Bidirectional- {LSTM}(BiLSTM), which is said to be one of the first benchmarks for different  {NMT}. The literature uses parallel sentence mining (PSM) as a data augmentation technique that identifies bitexts, which are translations of each other. In the end, the best system, which uses  {BPE}, BiLSTM and PSM, has reached a SacreBLEU score of 13.22.

\textbf{Transformer-based Neural Machine Translation}
 \citet{mak2022low} has developed a Transformer-based  {NMT} which outperforms Baidu Fanyi's Chinese-to-Cantonese Translation on 6 out of 8 test sets in BLEU score. It also has another major contribution in developing an effective approach to automatically extracting semantically similar sentences from parallel articles in Wikipedia, and it has obtained 72K parallel sentences.

These are some examples of literature that have explored Cantonese  {NMT}. Since the introduction of  {LLM}s, most related work which does not involve Transformers is likely outdated and will likely not produce state-of-the-art performance. Therefore this project has adopted a different route compared to most of them. However, there are some useful sources of datasets stemming from TransCan, and therefore, in the project, similar datasets have been used inspired by TransCan.

\section{\textsc{CantonMT} Methodology}
\label{cha:Method}
This section outlines the methodology used to achieve the paper's objectives. It began with a detailed procedure for collecting data for the upcoming model fine-tuning process with both parallel and monolingual corpus. It then discusses different baseline models, including hyper-parameters for fine-tuning and training strategies. After that, the back-translation procedure is discussed, including employing a model-switch strategy.
The design of the open-source web application can be found in Section \ref{app:toolkit}.
\subsection{Datasets and Preprocessing}
Several datasets are used in this paper for training and evaluation. Since Cantonese-English parallel corpora are not readily available, combinations of different datasets are used for the initial training of baseline models. Furthermore, to aid the back-translation strategy in the latter part of the project, monolingual corpora for both Cantonese and English are required, and therefore, they will be discussed in the following section.
\subsubsection{Parallel Corpus}
To fine-tune different baseline models, a parallel corpus is required to train the model to translate Cantonese to English at a reasonable level. 
In the end, three different parallel corpora are found between different timestamps of the investigation. Therefore, the latter two are used for training only, while the former are used for training and evaluation.

\begin{figure*}[ht]
\centering
\includegraphics[width=0.99\textwidth]{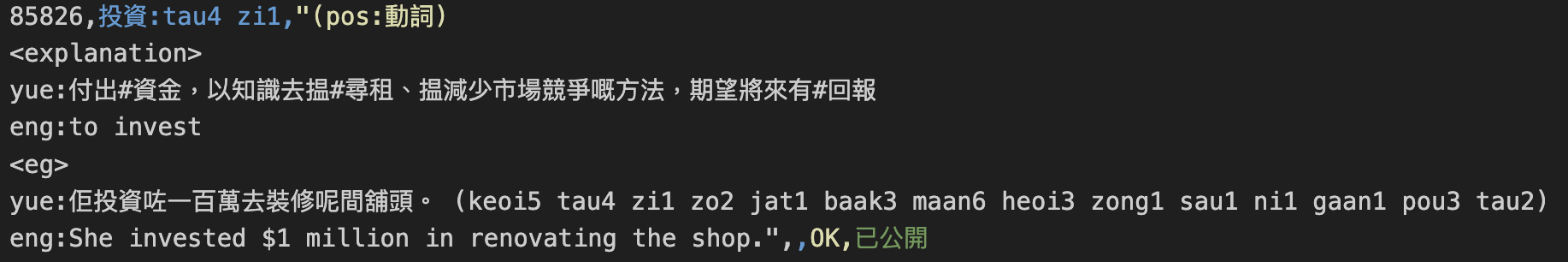}
    \caption{Sample Data Format for Words.hk}
    \label{fig:wordshk}
\end{figure*}

\textbf{Words.hk Corpus}
Words.hk\footnote{\url{https://words.hk}} is an open Cantonese-English dictionary publicly available for people to download. 
We used the full dataset from their website, which contains different Cantonese words and some example sentences with their English translation. An example of the word 
\begin{CJK*}{UTF8}{bkai}
``投資 / touzi''\end{CJK*}
 in the dictionary is given in Figure \ref{fig:wordshk}.

\begin{figure*}[ht]
\centering
\includegraphics[width=0.5\textwidth]{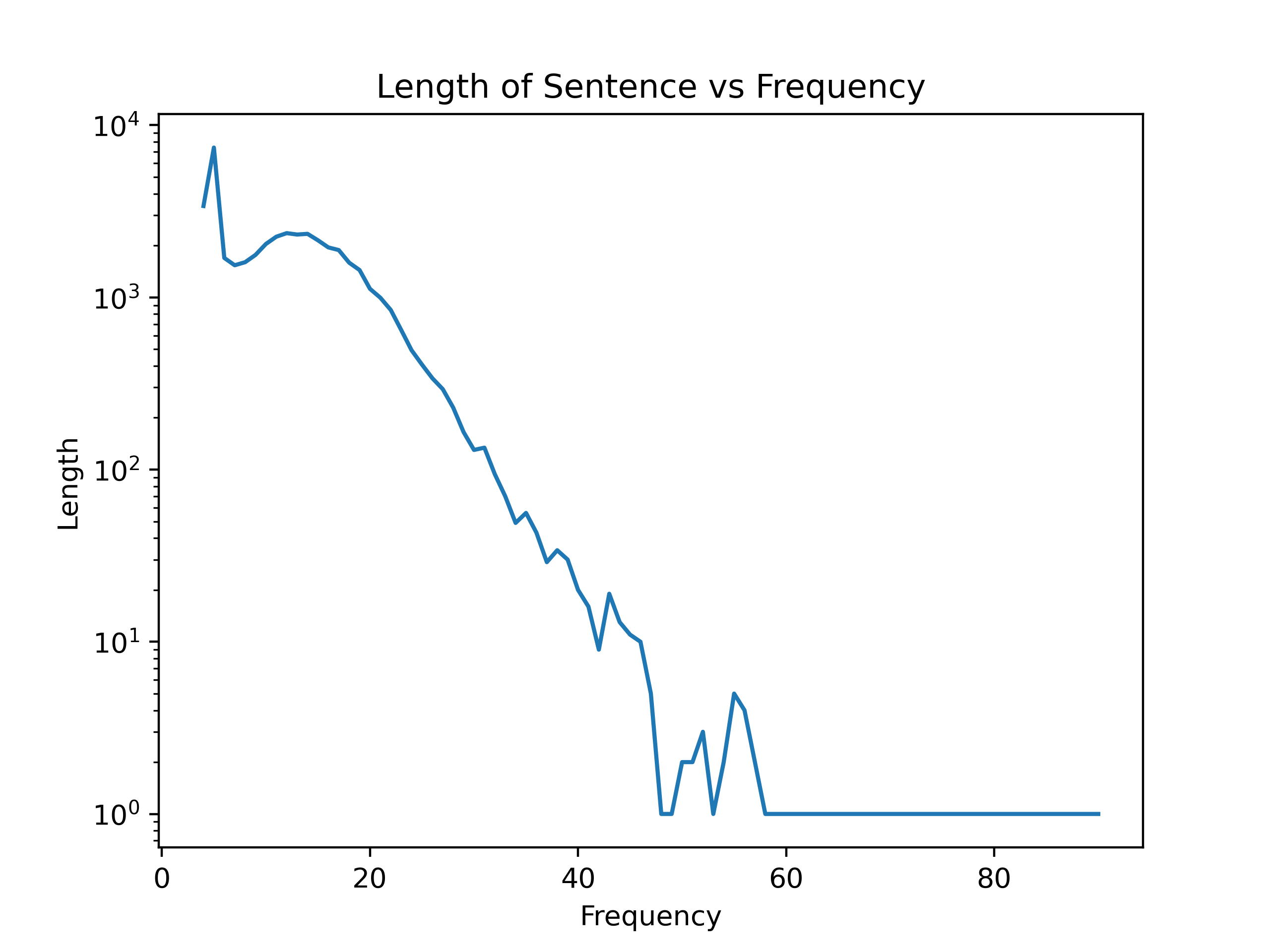}
    \caption{Words.hk - Sentence Length}
    \label{fig:wordshk-sentence}
\end{figure*}

From the data, only the sentence after the tag \textit{eng} has been used in this case, the sentence, ``\textit{She invested \$1 million in renovating the shop}'', has been extracted and also its corresponding Cantonese translation which is the sentence after the tag \textit{yue}. Data pre-processing has also been done, including removing hashtags and space since there is quite a lot in the dataset, potentially affecting data quality. In addition, there are sentences with multiple translations; in that case, the first translation has been taken. In the end, 44K sentences have been extracted from the dataset. A graph of the frequencies of the length of the Cantonese sentence has been plotted in Figure \ref{fig:wordshk-sentence}.\\

It is noticed that despite the effort only to keep sentences and no definitions, there are still quite a lot of short sentences in the dataset. Since for short sentences, it could be straightforward for the model to translate and, therefore, may lead to a bias in the evaluation, we have decided to split the dataset into short sentences and long sentences, where short sentences are sentences that have ten characters or less. In the end, there are 19.4K short sentences and 24.6K long sentences.\\

Since data are already very scarce, we have decided not to opt into the standard train-dev-test split of 8/1/1 or 7/2/1 and instead went for the approach of a 3K dev set and 3K test set. The reason behind this is based on that the standard practice for Workshop of Machine Translation (WMT)\footnote{\url{https://www2.statmt.org/wmt24}} shared task uses around 3K sentences for test sets when comparing different  {MT} systems.

\textbf{Wenlin Corpus}
Wenlin Institute \footnote{\url{https://wenlin.com}} creates software and dictionaries for learning the Chinese language, and there is a dictionary, ABC Cantonese-English Comprehensive Dictionary, which is readily available for registered users to use for research purposes. The process to obtain the dataset, however, is not straightforward. It involves first getting a list of URLs which store the data, and after that, it requires web scraping; at the end, an XML file is obtained, which includes all the sentences and other content.

Extracting is required to convert an XML file to a parallel corpus after obtaining an XML file. Based on initial inspection, the sentence should be inside the tag \textit{WL}; therefore, regular expression techniques are used to extract those sentences. After that, similar pre-processing as Words.hk has been done to obtain the training set and 14.5K parallel sentences are extracted.

\textbf{Opus Corpora}
Opus Corpora  \citep{tiedemann2004opus} is a collection of translated documents collected from the internet. The corpus is already aligned, and therefore, no pre-processing is required. It can be easily downloaded via their website \footnote{\url{https://opus.nlpl.eu/}}. An additional 9.6K parallel sentences are added to the final training set.
A figure summarising the size of the parallel dataset is shown in Table \ref{tab:data}

\begin{table}[!ht]
    \centering
    \begin{tabular}{l|l}
        Source & Size \\ \hline
        Wenlin & 14476 \\ 
        Words.hk & 44045 \\ 
        Opus & 9588 \\ \hline
        Total & 68109 \\ 
    \end{tabular}
    \caption{Parallel Corpus Size}
    \label{tab:data}
\end{table}

\subsubsection{Monolingual Corpus}
To aid the process of back-translation, a monolingual corpus from both the source and target language is required to investigate the \textit{iterative back-translation} approach.

\textbf{English Corpus}
There are many English monolingual corpora available, and in this project, the dataset we have decided to use is from the WMT 2012 News Collection \citep{callison-burch-etal-2012-findings}. It can be downloaded on the WMT website and contains 434K sentences, which is more than required for the back-translation.

\textbf{Cantonese Corpus}
However, for the Cantonese corpus, it is difficult to find an existing monolingual corpus. There is a Hong Kong Cantonese Corpus (HKCanCor) available  \citep{lee2022pycantonese}. However, this is based on spontaneous speech and radio programs from the late 1990s and, therefore, might be outdated and there is the language evolution factors with time passing by. Another reason for not choosing the data is that it only consists of 10K sentences, which is insufficient for back-translation purposes. 

Based on findings from  \citet{liang2021covid}, there should be abundant data on social media, including Facebook, YouTube, Instagram and different local forums. Since it will be hard to filter out Hong Kong users who use Cantonese in their social media comments, we have decided to turn to local forums. There are few mainstream ones which have an abundance of data, including Baby-Kingdom\footnote{\url{https://www.baby-kingdom.com/forum.php}}, DiscussHK\footnote{\url{https://www.discuss.com.hk/}}, and LIHKG\footnote{\url{https://lihkg.com}}.

\begin{figure*}[ht]
\centering
\includegraphics[width=0.99\textwidth]{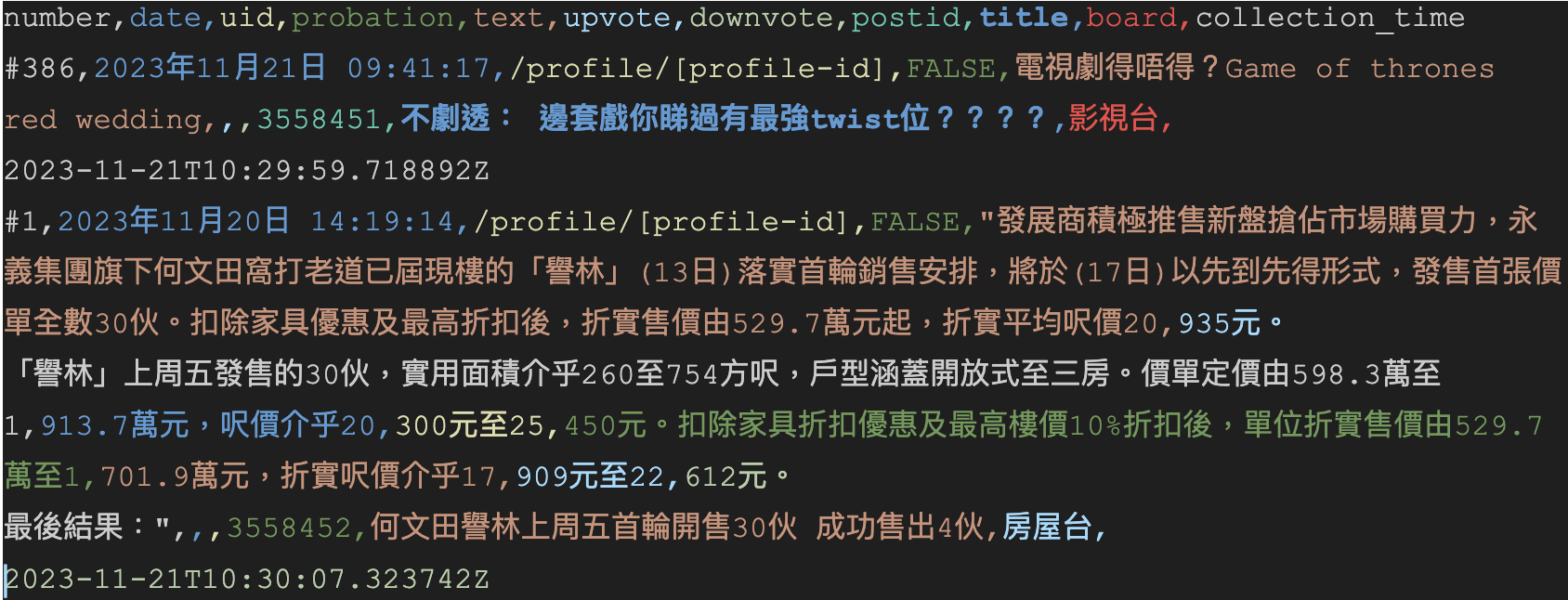}
    \caption{LIHKG Data Example}
    \label{fig:lihkg-data}
\end{figure*}

In the end, based on tools available online, we have decided to collect data from LIHKG. It is an online forum platform that was launched in 2016 and has multiple categories, including sports, entertainment,
hot topics, gossip, current affairs, etc. There is a scraper readily available online from  \citet{ho2020lihkgr}, which we have used to scrape the data from LIHKG. Data is scraped in CSV format, where an example can be seen in Figure \ref{fig:lihkg-data} (profile ID masked).

29K posts have been scraped, and only the text part has been used as the monolingual data. Some more \textbf{pre-processing} has been done to the data, including stripping all the links in the data and filtering out all the sentences shorter than 10 Chinese characters. In the end, 1.1M sentences have been scraped, which is more than enough for our investigation. 
We \textbf{shuffled} the dataset so that it can be used by the research community for free, as long as they sign a user agreement form for non-commercial usage. 

\subsection{Baseline Models}
\subsubsection{Model Selections}
In this training phase, we aim to train a set of reasonable Cantonese-English  {MT} models for model comparisons as baselines and synthetic data generation. Three different models have been chosen for this project, including Opus-MT, NLLB and mBART. Model information can be seen in Table \ref{tab:models-parameter}

\begin{table}[!ht]
    \centering
    \begin{tabular}{llll}
        \toprule
        ~ & Opus & NLLB & mBart \\ \hline
        Layers & 12 & 24 & 24 \\
        Hidden Unit & 512 & 1024 & 1024 \\
        Model Parameters & 77.9M & 615M & 610.9M \\
        Language Pair & No & Yes & No \\
        Release Year & 2020 & 2022 & 2020 \\ \bottomrule
    \end{tabular}
    \caption{Parameters from deployed models.
    Language pair: if the model contains Cantonese-English as a language pair}
    \label{tab:models-parameter}
\end{table}

\textbf{Pivot Model}
NLLB, as mentioned in Section \ref{cha:tech-bg}, is the only  {LLM} that states that it supports Cantonese  {MT} and, therefore, is naturally chosen as a pivot model. Further evaluation also supports the selection. With NLLB as a pivot model, two further investigations will be made using the rationales below.

\textbf{Model Size}
\textit{Does model size affect translation performance?} Since the version of NLLB we have chosen contains 600M parameters, a much smaller model was chosen to investigate this question. Therefore, Opus-MT was selected because it includes 78M parameters, around an 8x difference in size.

\textbf{Pre-training with the targeted language}
\textit{How much does it matter if the pre-trained translation models have Cantonese in their pre-training?} For this, mBART has been chosen, another  {LLM} which should not have Cantonese as pre-training or far less when compared to NLLB. The model also contains a similar number of parameters, and therefore, pre-training with the investigated language (Cantonese) should be the only factor.

\subsubsection{Fine-Tuning}
During this phase, only Words.hk data were discovered; therefore, the initial fine-tuning was conducted on this dataset.

The fine-tuning process has mostly been done on the Hugging Face\footnote{\url{https://huggingface.co}} API since all the models can be found on their website. Training has been done with Adam Optimizer \citep{kingma2017adam} with an initial learning rate of $10^{-4}$. 
In terms of Epoch, since fine-tuning the different models takes different ranges of time and similar training time is hoped to obtain a fair result, 10 epochs are used for the OPUS-MT model, and 3 epochs are used for NLLB and mBART. 

\subsection{Back-Translation}
After fine-tuning the baseline models, back-translation can be done where an iterative approach is first tried since both source and target language monolingual corpus are present. 
The following approach is inspired by the literature from  \citet{hoang2018iterative} with some changes. A general outline of the iterative approach is given here.

\begin{enumerate}
  \item Train a Baseline Model FM\textsc{0} for the forward direction (Cantonese to English)
  \item Generate synthetic data CANTO-SYN\textsc{1} with Cantonese monolingual data and model FM\textsc{0}
  \item Train different synthetic model BM\textsc{1} with different ratios of gold standard data and synthetic Data CANTO-SYN\textsc{1}
  \item Select best model from BM\textsc{1} with SacreBLEU and Generate synthetic data ENG-SYN\textsc{1} with English monolingual data and model BM\textsc{1}
  \item Train different synthetic model FM\textsc{1} with different ratios of gold standard data and synthetic Data ENG-SYN\textsc{1}
  \item Repeat Step 2-5
\end{enumerate}

\begin{figure*}[ht]
\centering
\includegraphics[width=0.99\textwidth]{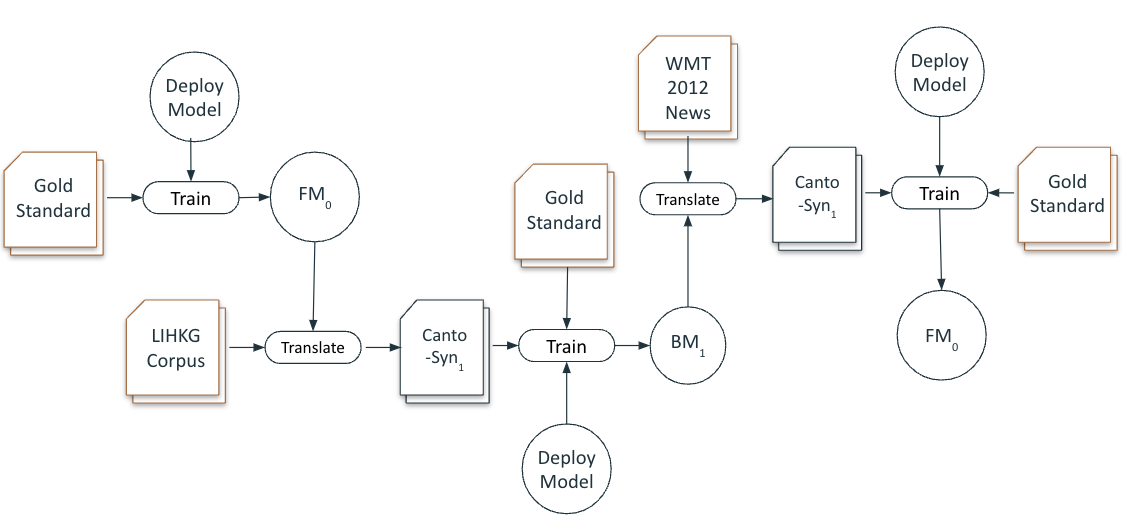}
    \caption{Diagram for Iterative Back Translation}
    \label{fig:IBT}
\end{figure*}

Each step will now be explained in a more detailed manner, and a diagram is provided to understand the step visually in Figure \ref{fig:IBT}.\\

\textbf{Train Baseline Model}
This was done in the previous phase. Since our main focus will be on NLLB, the model trained throughout this section will be NLLB.

\textbf{Generate Synthetic Parallel Data from Cantonese Data}
We can now use the fine-tuned model to generate a synthetic parallel corpus. we have generated 200K synthetic data, randomly sampled in the LIHKG corpus using the baseline model from the previous phase. 200K is chosen since it is 5x larger than the gold standard data, which is the maximum ratio with which we will be diluting the gold standard data.

\textbf{Fine-tuning Backward Model with Synthetic Data}
After generating synthetic data, we can now start fine-tuning the backward direction model. Inspired by  \citet{hoang2018iterative}, where different ratios of synthetic data and real data are used, a similar approach has also been made here. We have experimented with ratios of 1:1, 1:2, 1:3 and 1:5 for the backward model, where all the gold standard data have been used and added with the correct ratio of synthetic data randomly sampled from the previously generated corpus. 
All models are trained from the deployed version of NLLB.

\textbf{Generate Synthetic Parallel Data from English Data}
After training a set of models, the best model is chosen based on the best SacreBLEU score with the development set of gold standard data. Another set of synthetic data will then be generated with the WMT2012News corpus and the best model chosen from the last part. 200K sentences have been randomly sampled and put into the model for translation in preparation for a new synthetic data set.

\textbf{Fine-tuning Forward Model with Synthetic Data}
The generated synthetic data can now be used to fine-tune the forward model, where the same setup is used compared to the previous fine-tuning of the backward model. An additional ratio is also experimented with by \textit{sampling half of the gold standard data and pairing it with the same number of synthetic sentence pairs}.

\textbf{Limitations}
Due to the lack of computing power and the fact that the best backward model was the one with no synthetic data, no iterations were made afterwards.

\subsection{Plug-and-Play: Model-Switch Mechanism}
With different models available at our disposal, another way of experimenting with back-translation has been thought of, where a different type of model will be used in the opposite direction, which we call the method plug-and-play model or more formally, Model Switch. The idea goes through a process similar to the last section, where different models are chosen instead of choosing NLLB as a model throughout the iteration.

Since NLLB, in theory, should be the best model (most knowledge), a decision has been made to include NLLB at least in 1 direction. Below is the list of model pairs that are experimented with:
\begin{enumerate}
    \item Forward: NLLB, Backward: Opus-MT
    \item Forward: NLLB, Backward: mBART
    \item Forward: mBART, Backward: NLLB
    \item Forward: Opus-MT, Backward: NLLB
\end{enumerate}

One thing to note is that to balance out the training time for each model, the Opus-MT model is trained for 10 epochs instead of 3 epochs for the other two models.

\section{Experimental Evaluations}
\label{cha:experiments}
This section first covers the different evaluation processes used for the project, and the results are then reported and analysed in the latter part of the section.

\subsection{Evaluation Method}
After all the models have been trained, each model has been evaluated with a set of automatic metrics, where the best models are selected and compared against state-of-the-art translators.
We first introduce the automatic evaluation; then, it comes to the human evaluation settings where the modified (simplified) HOPE metric -- HOPES -- will be used (Section \ref{Eval_Sec_HumanEval}).

\subsubsection{Comparisons to the State-of-the-Art 
}
During the evaluation, state-of-the-art translators are also included to compare against other models, such as Bing and Baidu. The translation is obtained with their API using their corresponding Cantonese and English language pair. Furthermore, 
GPT-4 are also included for comparisons where a fine-tuned version towards translation to Cantonese by ``Community Builder" called Cantonese Companion \footnote{ \url{https://chat.openai.com/share/7ee588af-dc48-4406-95f4-0471e1fb70a8}} is used.
\subsubsection{Automatic Evaluation}

\textit{{Lexicon-based Metrics: SacreBLEU and hLEPOR}}.
For {MT} task, one of the most used metrics when evaluating {MT} system translation performance is BLEU  \citep{papineni2002bleu}. However, since BLEU could vary in different literature, \textbf{SacreBLEU}  \citep{post2018sacrebleu} is developed in the hope of having a standard way of calculating the BLEU score. It is, therefore, used as the main metric throughout the project as an evaluation during training.

However, it should be said that with only one automatic metric, especially with a variant of BLEU, it might not fully capture and understand which is the best model. \citep{callison2006evaluating}. Therefore, another lexical-based metric is used.
\textbf{hLEPOR}  \citep{han-etal-2013-language} is another lexical-based metric which has reported much higher correlation scores to the human evaluation than BLEU and other lexical-based metrics on the WMT shared task data  \citep{han2013hlepor}. Therefore, it is used to pair up with SacreBLEU.

\textit{{Neural-based Metrics: COMET and BERTscore}}.
In the recent WMT findings, 
it is reported by  \citet{freitag2022results} that neural-based metrics are significantly better than non-neural metrics regarding the correlation of human evaluation. 
Therefore, an additional two metrics have been used, i.e. \textbf{COMET} and \textbf{BERTscore}, of which COMET-22 was ranked the second highest metric from the WMT shared task, and BERTscore is used commonly in the {MT} community nowadays.

{\textbf{Implementations}}
Each trained model and state-of-the-art translator will translate 3K sentences in the test set. Since all the metrics are reference-based, the gold standard translation is required, and therefore, both the translation and reference are passed on for the metric calculation. For COMET-22, SacreBLEU and BERTScore, Hugging Face Evaluate\footnote{\url{https://huggingface.co/docs/evaluate/index}} modules are used to support the calculation, where hLEPOR has its own Python module\footnote{\url{https://pypi.org/project/hLepor/}} for its calculation. The results are reported in the next section.

\subsubsection{Human Evaluation}
\label{Eval_Sec_HumanEval}
Even with four different automatic metrics, it is still hard to judge the model's performance based on those chosen metrics. Therefore, human evaluations are also being conducted to understand better the comparison with state-of-the-art models and the \textit{different types} of errors that the trained models or deployed translators tend to make.

\textbf{{HOPES framework}
}With that in mind, we have borrowed the HOPE framework  \citep{gladkoff-han-2022-hope}. The original HOPE framework includes eight detailed error types from industrial practice.
However, upon our review, some error types can be merged to make the human evaluation task more efficient and better match our data, where a modified framework, HOPE-Simplified (HOPES), is proposed. The merging procedure is shown in the below list.
\begin{enumerate}
    \item \textbf{Merge Impact(IMP) and Mistranslation(MIS) as MIS}:\\
    The definitions of IMP and MIS are ``The translation fails to convert main thoughts clearly'' and ``Translation distorts the meaning of the source and presents mistranslation or accuracy error'' respectively. They overlap in accuracy and meaning preservation from the source sentence, which both reflect the semantics error. \\
    Therefore, it is merged as Mistranslation(MIS), where the new definition is given as ``perceived meaning differs from the actual meaning''.
    
    \item \textbf{Merge Terminology(TRM) and Proper Name (PRN) as Terms(TRM)}:\\
    The original definitions of TRM and PRN are ``incorrect terminology, inconsistency on the translation of entities'' and ``a proper name is translated incorrectly'' respectively.
    In our experimental data, the name is not popular, and proper names can be entity types if they appear in the test set.\\
    Furthermore, the original data does not define the scoring mechanism in a specific way. For example, when the translation mistranslates a critical word, should it be given as a critical error since it distorts the meaning, or a minor error since there is only one mistake in the translation? With the newly defined MIS, the first case could be covered by that, and therefore, a minor error should be given. \\
    Therefore, the error types are merged as TRM, with the new definition of ``Incorrect terminology'', including proper names or inconsistency of translation of entities, where a higher score means there are more incorrect terms''.
    
    \item \textbf{Merge Style (STL), Proofreading (PRF), Required Adaptation Missing (RAM) into Style(STL).}\\
    The original definitions of these three are ``translation has poor style but is not necessarily ungrammatical or formally incorrect'', ``linguistic error which does not affect accuracy or meaning transfer but needs to be fixed'', and ``source contains error that has to be corrected or target market requires substantial adaptation of the source, which translator failed to make; impact on the end user suffers''.
    These errors are all related to localisation and adaptation.
    We \textit{summarise} the merged error type Style as ``Translation has poor style, but is not necessarily ungrammatically or formally incorrect. It may also include linguistic error which does not affect meaning, but potentially makes the end user suffer''.
\end{enumerate}

There is also a human annotation guideline in Appendix \ref{appendix:hopes-guide}, in which examples of critical errors for each mistake and the scoring guide are given to each annotator.
Based on literature from  \citet{gladkoff2022measuring} regarding evaluation uncertainty, less than 200 human evaluation sentences are insufficient to make a statistical significance. Therefore, 200 sentences from the test set are randomly sampled from the test set and used for human evaluation.

Three different translation systems are chosen, including the best model from our training, one of the commercial translators and community-finetuned GPT4. The reasons are based on the evaluation metric scores, which will be shown in the latter part.

There were a total of 4 annotators who are fluent English speakers and native Cantonese users annotated the translations for the 200 x 3 translations. Each translation is then evaluated by two annotators to measure the agreement level between them, and therefore, the results should be more accurate and reflect the performance of each system. It should also be noted that the results can also help us understand the general error types the models are making, which may be useful for future work.

\begin{table*}[!htb]
    \centering
    \begin{tabular}{l|cccc}
        \toprule
        Model Name & SacreBLEU & hLEPOR & BERTscore & COMET \\ \hline
        nllb-forward-bl & \textbf{16.5117} & \textbf{0.5651} & \textbf{0.9248} & \textbf{0.7376} \\
        mbart-forward-bl & 15.7513 & 0.5623 & 0.9227 & 0.7314 \\
        opus-forward-bl-10E & 15.0602 & 0.5581 & 0.9219 & 0.7193 \\
        \hline\hline 
        nllb-200-deploy-no-finetune & 11.1827 & 0.4925 & 0.9129 & 0.6863 \\
        opus-deploy-no-finetune & 10.4035 & 0.4773 & 0.9082 & 0.6584 \\
        mbart-deploy-no-finetune & 8.3157 & 0.4387 & 0.9005 & 0.6273 \\
        \hline
        \bottomrule
    \end{tabular}
    \caption{Evaluation Scores for Baseline(bl) Models and deployed model, where NLLB and mBART are trained with three epochs, and Opus is trained with ten epochs (10E) }
    \label{tab:bl-eval}
\end{table*}

\begin{figure*}[ht]
\centering
     \begin{subfigure}{0.45\textwidth}
         \centering
         \includegraphics[width=\textwidth]{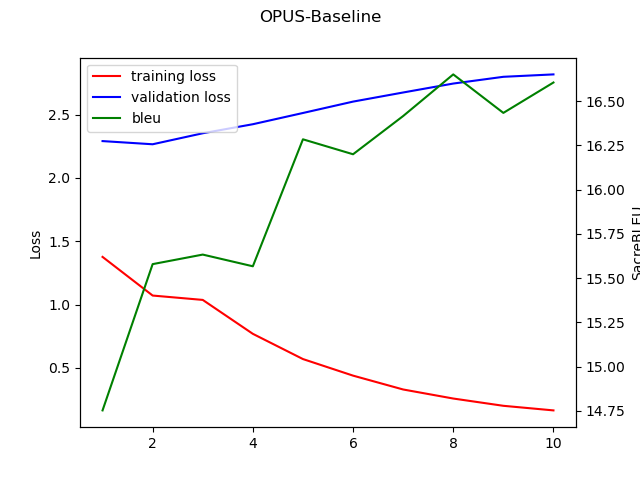}
         \caption{OPUS-MT Training Curve - 10E}
         \label{fig:opus-bl-train}
     \end{subfigure}
     \hfill
     \begin{subfigure}{0.45\textwidth}
         \centering
         \includegraphics[width=\textwidth]{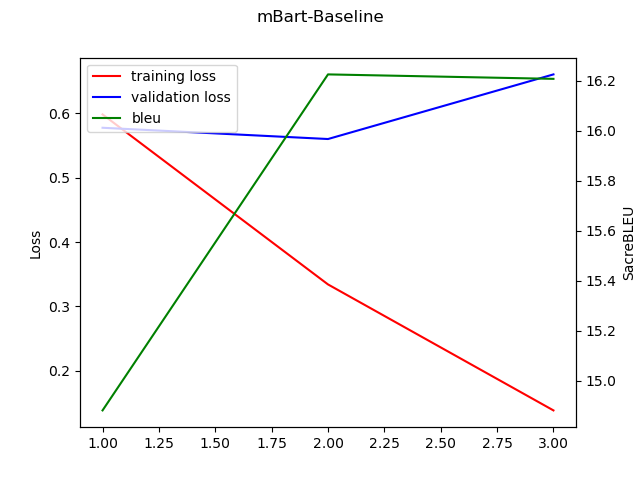}
         \caption{mBART Training Curve - 3E}
         \label{fig:mbart-bl-train}
     \end{subfigure}
     \begin{subfigure}{0.45\textwidth}
         \centering
         \includegraphics[width=\textwidth]{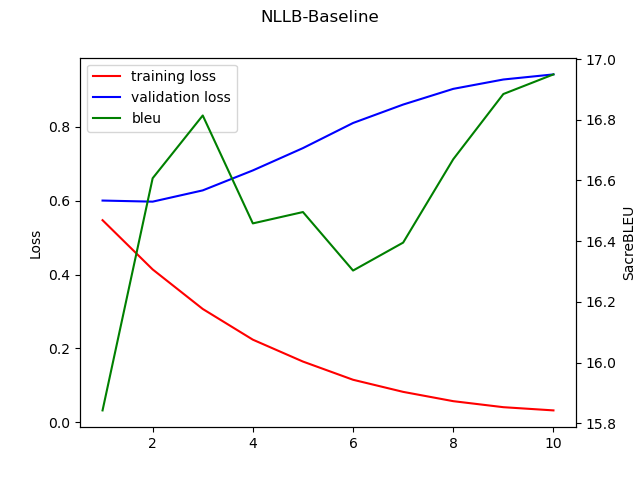}
         \caption{NLLB Training Curve - 10E}
         \label{fig:nllb-bl-train}
     \end{subfigure}
        \caption{Training Curves for Various Models}
    \label{fig:training-curves}
\end{figure*}
\subsection{Baseline Models}
The results for the fine-tuned baseline model are shown in Table \ref{tab:bl-eval}. Additionally, the learning curves for each baseline model have been drawn and shown in Figure \ref{fig:training-curves}.
The learning curves show that the NLLB model reaches its highest score at Epoch 3, followed by a significant decline until Epoch 6, after which it recovers at Epoch 10. Therefore, using only three epochs on training with the rest of the models is justifiable. On the other hand, the Opus-MT model exhibits a gradual improvement in the SacreBLEU score across epochs despite experiencing minor fluctuations along the way.

It can also be seen from the results that NLLB scores the highest across all four metrics, and therefore, it can be concluded that it performs the best for the translation task, which further justifies the choice of the pivot model. Furthermore, it can also be said that larger models and models with Cantonese in the pre-training produce better outputs. 

Another result that can be seen is that with minimal fine-tuning, all three baseline models have a much higher score when compared to the deployed models, improving from 5 to 7 in terms of absolute SacreBLEU score; the same phenomenon can also be seen in other metrics.

\begin{figure*}[]
\centering
\includegraphics[width=0.5\textwidth]{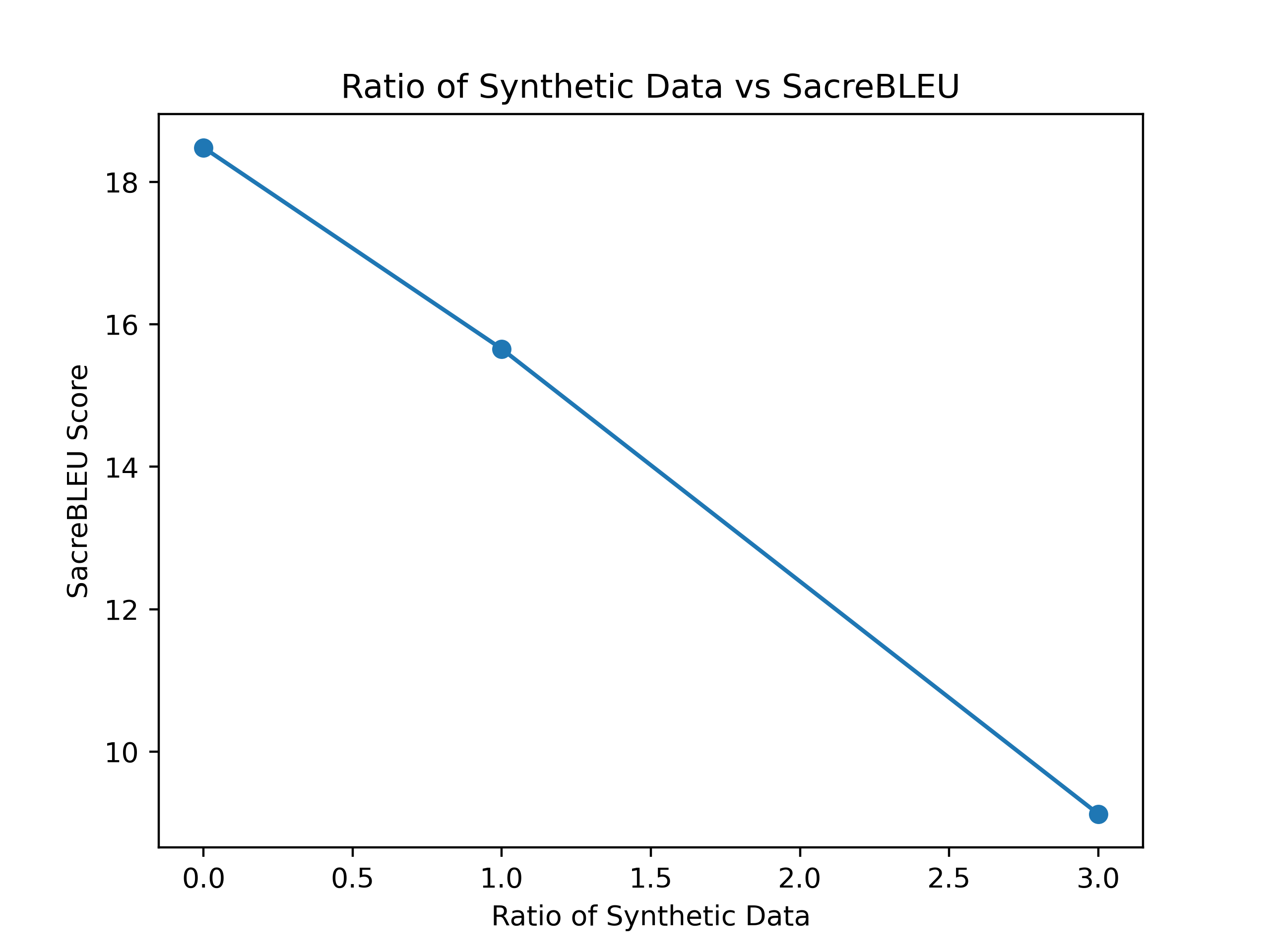}
    \caption{Effect on SacreBLEU with different ratios of synthetic data}
    \label{fig:back-syn}
\end{figure*}

\subsection{Back-Translation}

\begin{table*}[!htb]
    \centering
    \begin{tabular}{l|cccc}
        \toprule
        Model Name & SacreBLEU & hLEPOR & BERTscore & COMET \\ \hline
        nllb-forward-bl & 16.5117 & 0.5651 & 0.9248 & 0.7376 \\
        nllb-forward-syn-h:h & 15.7751 & 0.5616 & 0.9235 & 0.7342 \\ 
        nllb-forward-syn-1:1 & \textbf{16.5901} & 0.5686 & \textbf{0.925} & \textbf{0.7409} \\ 
        nllb-forward-syn-1:1-10E & 16.5203 & \textbf{0.5689} & 0.9247 & 0.738 \\ 
        nllb-forward-syn-1:3 & 15.9175 & 0.5626 & 0.924 & 0.7376 \\ 
        nllb-forward-syn-1:5 & 15.8074 & 0.562 & 0.9237 & 0.7386 \\ \hline
        \bottomrule
    \end{tabular}
    \caption{Evaluation Scores for Baseline(bl) Models and model trained with synthetic data (syn), where the first number indicates the ratio of the gold standard data used (h=half), and the second number indicates the ratio of synthetic data used}
    \label{tab:BT-eval}
\end{table*}

\begin{figure*}[]
\centering
\includegraphics[width=0.5\textwidth]{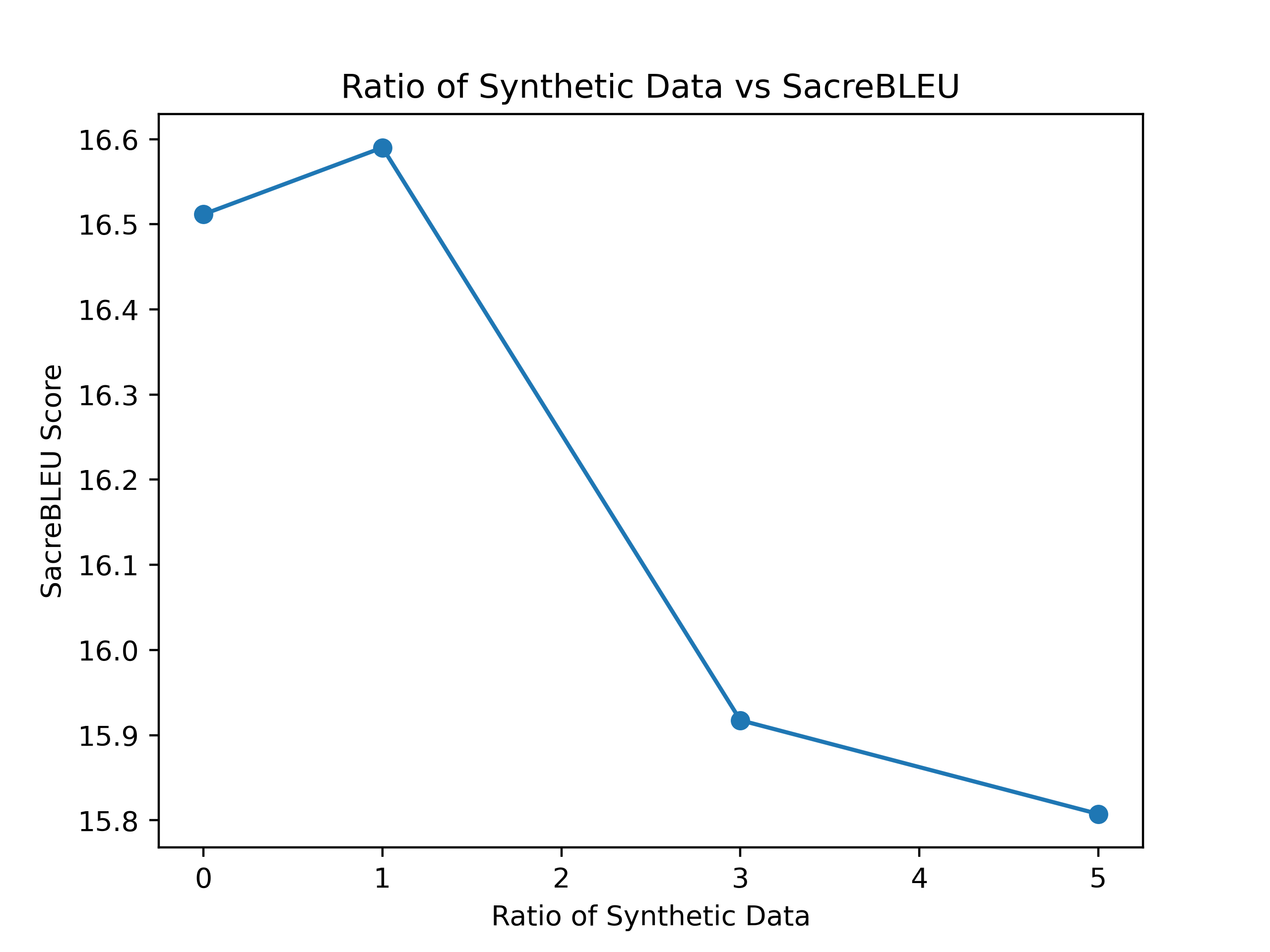}
    \caption{Effect on SacreBLEU with different ratios of synthetic data}
    \label{fig:forward-syn}
\end{figure*}
The first phase of back-translation involves training a backward model (English -> Cantonese), where synthetic data are used in a set of ratios. A graph showing the SacreBLEU score against the ratio used ranging from 0 to 3, where 0 indicates no synthetic data is used, is shown in Figure \ref{fig:back-syn}.
The results indicate that the quality of the translation drops significantly with synthetic data. Potential reasons that may lead to the result are poor translation quality for the forward model or poor quality of the monolingual corpus. We are more inclined toward the latter since back-translation has proven to improve the system a lot in the literature on other language pairs. A potential improvement in the future for this aspect can be cleaning the dataset with different methods, including grammar checking, which is proposed by  \citet{pham2023engviet}.

Therefore, the chosen model for generating back-translation synthetic data is the baseline model trained with no synthetic data where 200K sentences are generated through the model selected.
The results for back-translation with a single model - NLLB are shown in Table \ref{tab:BT-eval} and a graph showing the SacreBLEU score against the ratio used ranging from 0 to 5, where 0 indicates no synthetic data is used, is shown in Figure \ref{fig:forward-syn}.

Based on the results, NLLB-syn-1:1 slightly outperforms NLLB-bl across all metrics. However, increasing the proportion of synthetic data, as in 1:3 and 1:5 ratios, results in a decline of approximately one absolute SacreBLEU point. The results have shown that \textit{using synthetic data improves performance slightly}. Still, with too much synthetic data, since there might be more errors introduced to the data, the translation quality decreases. 

\begin{table*}[!htb]
    \centering
    \begin{tabular}{l|cccc}
        \toprule
        Model Name & SacreBLEU & hLEPOR & BERTscore & COMET \\ \hline
        nllb-forward-bl & 16.5117 & 0.5651 & 0.9248 & 0.7376 \\
        nllb-forward-syn-h:h & 15.7751 & 0.5616 & 0.9235 & 0.7342 \\ 
        nllb-forward-syn-1:1 & \textbf{16.5901} & 0.5686 & \textbf{0.925} & \textbf{0.7409} \\ 
        nllb-forward-syn-1:1-10E & 16.5203 & \textbf{0.5689} & 0.9247 & 0.738 \\ 
        nllb-forward-syn-1:3 & 15.9175 & 0.5626 & 0.924 & 0.7376 \\ 
        nllb-forward-syn-1:5 & 15.8074 & 0.562 & 0.9237 & 0.7386 \\ \hline
        nllb-forward-syn-1:1-mbart & \textbf{16.8077} & \textbf{0.571} & \textbf{0.9256} & \textbf{0.7425} \\
        nllb-forward-syn-1:3-mbart & 15.8621 & 0.5617 & 0.9246 & 0.7384 \\
        nllb-forward-syn-1:1-opus & 16.5537 & 0.5704 & 0.9254 & 0.7416 \\
        nllb-forward-syn-1:3-opus & 15.9348 & 0.5651 & 0.9242 & 0.7374 \\ \hline
        mbart-forward-bl & 15.7513 & 0.5623 & 0.9227 & 0.7314 \\
        mbart-forward-syn-1:1-nllb & \textbf{16.0358} & \textbf{0.5681} & \textbf{0.9241} & \textbf{0.738} \\
        mbart-forward-syn-1:3-nllb & 15.326 & 0.5584 & 0.9225 & 0.7319 \\ \hline
        opus-forward-bl-10E & \textbf{15.0602} & \textbf{0.5581} & \textbf{0.9219} & \textbf{0.7193} \\
        opus-forward-syn-1:1-10E-nllb & 13.0623 & 0.5409 & 0.9164 & 0.6897 \\
        opus-forward-syn-1:3-10E-nllb & 13.3666 & 0.5442 & 0.9167 & 0.6957 \\ \hline
        \bottomrule
    \end{tabular}
    \caption{Automatic Evaluation Scores from Model-Switch Models and previous results, where the prefix model name indicates the model type, where there is a postfix model, it indicates the model that generated the synthetic data.
    }
    \label{tab:model-switch-eval}
\end{table*}

\subsection{Plug-and-Play Model Outcomes}

Based on the previous results and the suspected data quality issue, no re-back translations are being done here, and the generated synthetic data from English to Cantonese are using models fine-tuned with gold standard data only.
Results are then reported for the model-switch models and are shown in Table \ref{tab:model-switch-eval}.
From the results, a similar conclusion from above can be drawn from the mBART model too, where mBART-syn-1:1 also outperforms mBART-bl, but more ratios of synthetic data will reduce the evaluation scores such as 1:3.

Surprisingly, none of the synthetic models for Opus-MT outperforms Opus-bl, which might indicate that \textit{with small models, the error introduced from the synthetic data might confuse the model with its pre-training} and lead to a drop in translation quality.

\textit{Model-switching could produce a better result when compared to the single-model approach}. Firstly, the NLLB model that was fine-tuned using synthetic data from mBART achieved higher scores than when it was fine-tuned with synthetic data generated from itself with 0.2 SacreBLEU point. Secondly, when mBART was fine-tuned using synthetic data produced by NLLB, its performance surpassed fine-tuned with only bilingual real data. Thirdly, Opus-MT demonstrated a different performance pattern compared to the other two models under similar circumstances probably for the same reason mentioned above.

\begin{table*}[!htb]
    \centering
    \begin{tabular}{l|cccc}
        \toprule
        Model Name & SacreBLEU & hLEPOR & BERTscore & COMET \\ \hline
        nllb-forward-bl & 16.5117 & 0.5651 & 0.9248 & 0.7376 \\
        mbart-forward-bl & 15.7513 & 0.5623 & 0.9227 & 0.7314 \\
        opus-forward-bl-10E & 15.0602 & 0.5581 & 0.9219 & 0.7193 \\
        nllb-forward-syn-1:1-mbart & 16.8077 & 0.571 & {0.9256} & {0.7425} \\
        mbart-forward-syn-1:1-nllb & {16.0358} & {0.5681} & {0.9241} & {0.738} \\
        \hline\hline 
        nllb-forward-all3corpus & \textbf{16.9986} & \textbf{0.583} & \textbf{0.927} & \textbf{0.7549} \\
        nllb-forward-all3corpus-10E & 16.1749 & 0.5728 & 0.9254 & 0.7508 \\
        mbart-forward-all3corpus & 16.3204 & 0.5766 & 0.9253 & 0.7482 \\
        opus-forward-all3corpus-10E & 14.4699 & 0.5621 & 0.9191 & 0.7074 \\ \hline
        \bottomrule
    \end{tabular}
    \caption{Evaluation Scores with More Data fine-tuning and selection of previous results}
    \label{tab:moredata-eval}
\end{table*}

\begin{figure*}[h]
\centering
\includegraphics[width=0.5\textwidth]{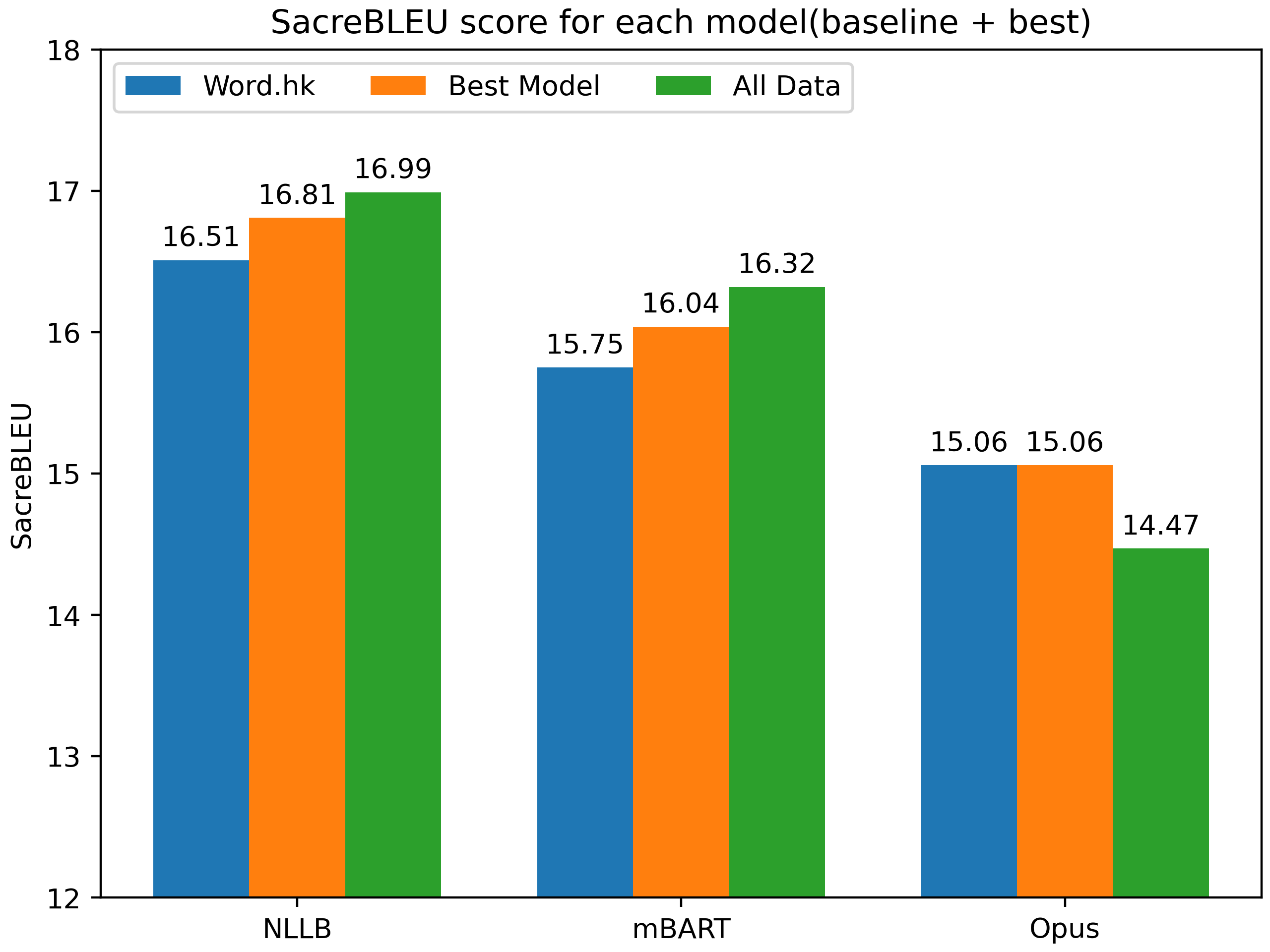}
    \caption{SacreBLEU score with the baseline model, best model and all three corpus fine-tuning}
    \label{fig:moredata-fig}
\end{figure*}

\subsection{Integrating More Real Data}
The additional collected datasets during the project are also used for fine-tuning the three different models, where results are shown in Table \ref{tab:moredata-eval} and Figure \ref{fig:moredata-fig}. From the results, it can be seen that fine-tuning with all three corpora outperforms the best model for NLLB and mBART; however, this is not the case for Opus-MT.
These results indicate the idea that Data Quality Matters, where even though less data is given (62K vs 76K), the models perform higher scores when compared to the best synthetic model from NLLB and mBART, respectively. This also could be a potential future work where the experiments can be re-conducted with more data, which should provide better results theoretically.

Another key point to this result is that the improvement shows that translation quality has generally improved, not just for the specific translation style in the test set. Since multiple datasets are combined for the training, the model should not be biased to the translation style in Words.hk but rather a more general style, which should provide a better translation output for human judgements, not just for automatic metric scores.

However, training with more data with OPUS-MT does not provide better performance, which might be suspect to the different translation styles in datasets; since it is a small model, there is a chance that it fully adapts to the Words.hk dataset rather than actually understanding the task of translation, where literature refers to this phenomenon as ``translationese''\footnote{Awkwardness or ungrammatically of translation, such as due to the overly literal translation of idioms or syntax.}, which may lead to high automatic metric score but low-quality translation.

Nevertheless, these outcomes demonstrated the possibility of improving model performances with more available real data, at least for NLLB and mBART models.

\begin{table*}[h]
    \centering
    \begin{tabular}{l|cccc}
        \toprule
        Model Name & SacreBLEU & hLEPOR & BERTscore & COMET \\ \hline
        nllb-forward-bl & 16.5117 & 0.5651 & 0.9248 & 0.7376 \\
        mbart-forward-bl & 15.7513 & 0.5623 & 0.9227 & 0.7314 \\
        opus-forward-bl-10E & 15.0602 & 0.5581 & 0.9219 & 0.7193 \\
        nllb-forward-syn-1:1-mbart & 16.8077 & 0.571 & {0.9256} & {0.7425} \\
        mbart-forward-syn-1:1-nllb & {16.0358} & {0.5681} & {0.9241} & {0.738} \\
        \hline\hline 
        baidu & 16.5669 & 0.5654 & 0.9243 & 0.7401 \\ 
        bing & 17.1098 & 0.5735 & 0.9258 & 0.7474 \\ 
        gpt4-ft(CantoneseCompanion) & \textbf{19.1622} & \textbf{0.5917} & \textbf{0.936} & \textbf{0.805} \\ \hline
        \bottomrule
    \end{tabular}
    \caption{Evaluation Score of State-of-the-art translators and selected models}
    \label{tab:sota-eval}
\end{table*}

\begin{figure*}[!htb]
\centering
\includegraphics[width=0.5\textwidth]{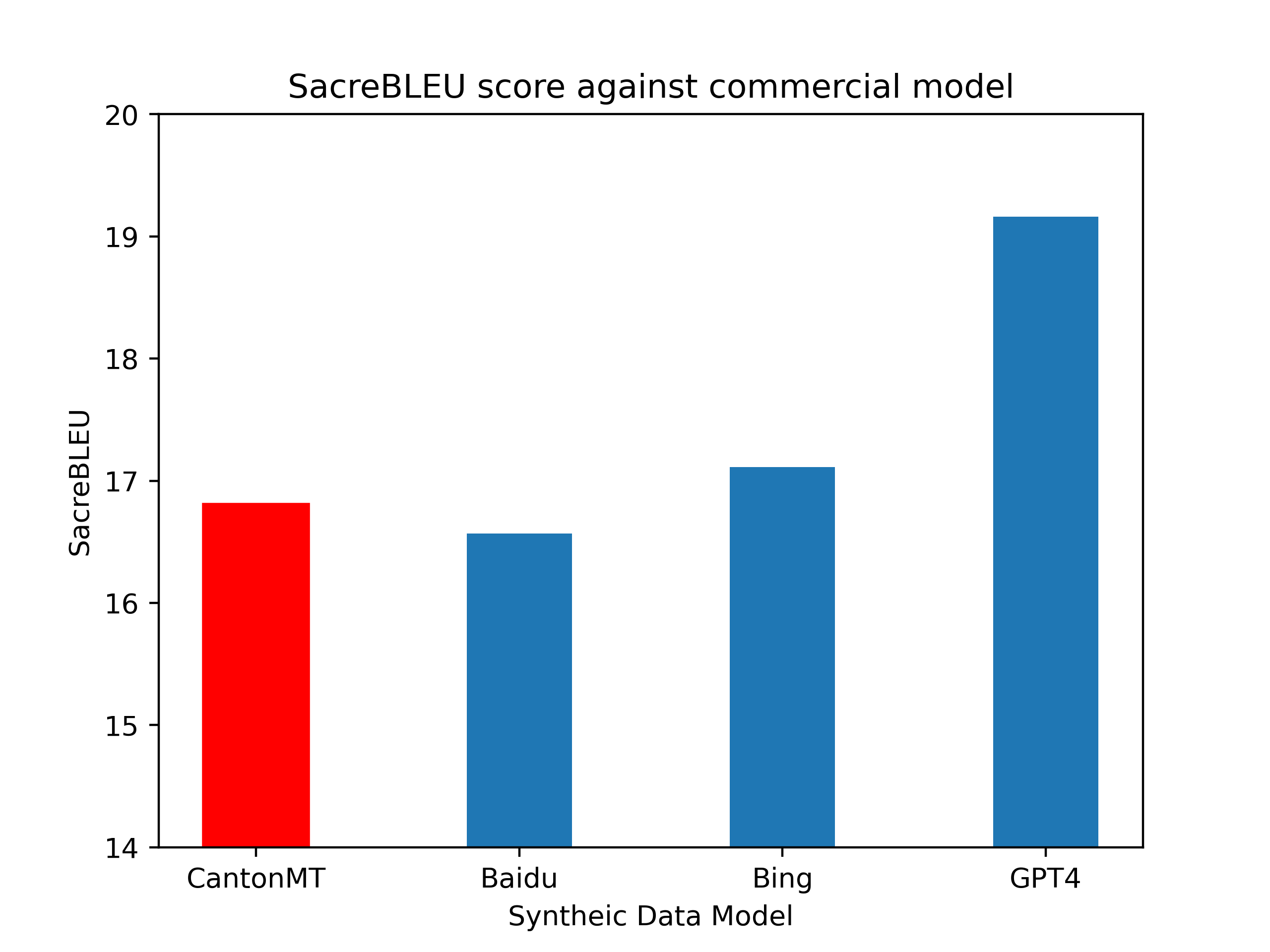}
    \caption{State-of-the-Art SacreBLEU Score against our best model}
    \label{fig:sota-fig}
\end{figure*}
\subsection{Comparisons to the State-of-the-Art }
State-of-the-art translators are then put into comparison with our models and results are shown in Table \ref{tab:sota-eval} and Figure \ref{fig:sota-fig}.

Cutting-edge {MT} technologies from commercial companies have shown varied performance, with GPT-4 fine-tuned models achieving the highest evaluation scores. However, this specific version of GPT is closed behind the paywall. Even with the free version, there are \textit{\textbf{limitations}} such as restrictions on the number of input tokens (\textit{length} of input) and a lack of \textit{transparency} regarding the underlying mechanisms of GPT-4's MT capabilities. Additionally, data \textbf{privacy} concerns arise when users opt for engines developed by commercial entities, e.g. when translating private texts such as the clinical domain as discussed by \citep{han2024neural}. 
On the other hand, our models stand out as an open-source alternative, offering users the flexibility to fine-tune the model with their data further or integrate additional models, ensuring full confidentiality for users.

In comparison, translators from Bing and Baidu have performed similarly to our best system. Notably, Bing's translator exhibits slightly superior performance over Baidu, particularly in lexical-based metrics like SacreBLEU and hLEPOR, highlighting its effectiveness in {MT} tasks.
More inspection of the specific translators against our best model can be seen in the next section, where human evaluation is conducted.
A table that combines all the automatic evaluation results can be seen in Appendix \ref{appendix:auto-eval}

\subsection{Human Evaluation}
Based on the previous evaluation results, three models/translators are chosen for human evaluation, which is GPT4, the best-performing translator; Bing, the best-performing commercial translator; and NLLB-syn-1:1-mBART, the best model\footnote{The extra datasets have not been found when human evaluation is conducted} in our system.

\subsubsection{{Text Degeneration}}
Upon first glance at the synthetic data and test set translations, some interesting phenomena are happening,  described as \textit{neural text degeneration}  \citep{Holtzman2020degeneration}. Examples of text degeneration can be seen in Table \ref{tab:text-degen}. From the example, ``handwritten'' has been repeated multiple times, indicating the models generate repetitive and dull loops. This could be another point of future work to adopt some methods for minimising these situations.\\
\begin{CJK*}{UTF8}{bkai}
\begin{table*}[!htb]
    \centering
    \begin{tabular}{|l|c|}
    \hline
         Source Sentence & 佢踢住對人字拖噉行出嚟。\\\hline
         Model Translation & He walked out with a pair of handwritten handwritten handwritten.\\\hline
    \end{tabular}
    \caption{Example of Text Degeneration}
    \label{tab:text-degen}
\end{table*}
\end{CJK*}

\subsubsection{Results}
The results are then used to calculate inter-annotator agreement (IAA), via a quadratic-weighted Cohen's Kappa metric  \citep{cohen1968weighted}, where the ratings are grouped into two individual raters. The results are shown in Table \ref{tab:cohen}.

\begin{table}[ht]
\centering

\begin{tabular}{l|cccc}
\toprule
Metric & NLLB & Bing & GPT4 \\ \midrule
MIS   & 0.6671 & 0.6102 & 0.5700 \\
TERM  & 0.5700 & 0.4775 & 0.3874 \\
STYLE & 0.1123 & 0.3490 & 0.0348 \\
GRAM  & 0.4212 & 0.2899 & 0.2850 \\ \hline
Overall  & 0.6230 & 0.6136 & 0.4935 \\ \hline\bottomrule
\end{tabular}
\caption{Cohen's Kappa for Different Models and Metrics}
\label{tab:cohen}
\end{table}

The results show that the annotators have a substantial agreement level in the category of mistranslation \citep{Landis1977cohen} and the overall rating, which is calculated by adding all 4 metrics together. For the other metrics, terminology and grammar have shown a moderate agreement between annotators. However, there seems to be a low agreement level for style, which suggests that the guidelines might need more refinement and detailed explanations, or more likely, translation style is very personal and should not be a major contributing factor to whether or not the translation is good or not.

Since the annotators have shown some kind of agreement, the results shown in Table \ref{tab:hopes-score} should have some indication of whether or not the translation is up-to-standard and can provide a better understanding of the models' performance. Another table can be seen in Table \ref{tab:hopes-error} for errors in individual models, where a major error is defined as a total score higher than 15 and a minor error is defined as lower than 15 but excluding 0. Translations with no errors in all 4 categories are defined as No error. 

\begin{table}[ht]
\centering
\begin{tabular}{l|cccc}
\toprule
Metric & NLLB & Bing & GPT4 \\ \midrule
MIS   & 4.8025 & 2.9875 & 0.7025 \\
TERM  & 3.62 & 2.1425 & 0.655 \\
STYLE & 3.01 & 2.3975 & 0.8425 \\
GRAM  & 1.1475 & 0.82 & 0.1575 \\ \hline
Overall  & 12.58 & 8.3475 & 2.3575 \\ \hline\bottomrule
\end{tabular}
\caption{Average Score for Different Models and Metrics}
\label{tab:hopes-score}
\end{table}

\begin{table}[ht]
\centering
\begin{tabular}{l|cccc}
\toprule
Errors & NLLB & Bing & GPT4 \\ \midrule
No Error   & 81 & 119  & 242\\
Minor Error  & 183 & 206 & 144 \\
Major Error  & 136 & 75 & 14 \\ \hline\bottomrule
\end{tabular}
\caption{Translation Errors for different models (200 sentences x 2 annotators for each model)}
\label{tab:hopes-error}
\end{table}

The results have shown that fine-tuned GPT4 ``CantoneseCompanion'' is by far the best model for translation, where over half of the translations have shown no errors, and only 3\% of translations have major errors according to the metric. Also, for the different metrics, GPT4 has shown similar performance except for grammar, which indicates that error types are quite diverse for GPT4.

Moreover, Bing performs better than the best model from NLLB, which is in line with the automatic metric. Nevertheless, both models have only around 25\% translation, which is error-free. In the evaluation, it can be seen that there are quite a few cases for both models to translate the sentence literally, which leads to some slang not being correctly translated and, therefore, affects the quality of translation.

For our system, most errors stem from either mistranslation or terminology, which is often correlated since when a term is not correctly translated, it often causes meaning loss in the sentence. It can also be noticed that most of the sentences are often grammatically correct, which should be expected since the decoder part of the Transformers is trained with large amounts of English data and, therefore, should be well-versed in grammar knowledge.

The result here shows that additional effort will be needed to surpass one of the commercial translators, where there should be more effort put into improving the model's knowledge of terminology and slang. For example, having a knowledge graph and knowledge base to represent different terminology and slang  \citep{zhao-etal-2020-knowledge,han-etal-2020-alphamwe} could potentially allow the model to understand more terminology in Cantonese. Further pre-training in Cantonese can potentially improve performance too. 
Some example translations from our model can also be seen in the Appendix (Table \ref{tab:example-output}).

\section{Conclusions and Future Work}
\label{cha:conclude}
This section combines the strategies and assessments covered, focusing on the central findings and contributions made to the {NLP} community. It acknowledges the limitations encountered along the way, setting the stage for subsequent efforts to enhance and expand upon the current milestones. Future work is proposed with the intent to capitalise on the progress made, address the shortcomings, and further research in this domain. 


\subsection{Findings}
The work focused on the investigation of data augmentation (via back- and forward-translation for synthetic data generation) and model-switch mechanism for Cantonese-to-English NMT, along with an open-sourced \textsc{CantonMT} toolkit release as well as the collected corpora.
The 4 main objectives that were outlined in the first section have been achieved:
\begin{itemize}
    \item \textbf{{Dataset Creation and Release}} A new parallel dataset has been created by merging various existing corpora, which could serve as a valuable resource for researchers in the Cantonese Natural Language Processing (NLP) community for further exploration in this field.
    Additionally, a substantial Cantonese monolingual dataset has been scraped and processed from an online forum, including anonymisation and data cleaning. This dataset could be extremely valuable for future research, such as Cantonese word embedding training for downstream NLP tasks, especially since, to the best of our knowledge, the data from the online forum is not publicly available.
    \item \textbf{Model Investigations: novel methods for Cantonese-English translation} Several models are developed and trained via back-translation to generate a synthetic parallel corpus for fine-tuning purposes, and the best models have reached comparable results against State-of-the-art commercial translators including Microsoft Bing and Baidu Translators, on the ground that we have very limited resources both computation and corpus wise. This work is the first to apply \textit{back-translation generated synthetic data} and \textit{model switch mechanisms} for augmenting Cantonese-English NMT in the field.
    \item \textbf{Experimental Evaluations: both automatic and human evaluations} Extensive experiments were conducted to evaluate the performances from trained models and the off-the-shelf translators, where multiple automatic evaluation metrics were used for comparison purposes, from both lexicon-based (BLER, hLEPOR) and embedding-space (COMET, BERTscore) categories. Furthermore, a standard human evaluation framework (HOPE)  \cite{gladkoff-han-2022-hope} has been modified and customised in this paper to compare different state-of-the-art translators against our best systems, where a clearer image of the models' strengths and weaknesses can be seen.
    \item \textbf{\textsc{CantonMT} User Interface} A highly modular full-stack web application has been designed and developed as an open-sourced translation platform that can act as a toolkit for other researchers to add different models and language pairs to our platform and software packages. These have achieved the designed objectives of this investigation, where an open-sourced Cantonese-to-English translation tool has been developed. 
\end{itemize}



\subsection{Challenges and Limitations}
We list some limitations of our current investigation.
\begin{itemize}
    \item \textbf{{Data Restrictions}} Even with a certain amount of data found during the investigation, we are aware that merely 60K data is not enough to train a model to reach expert-level performance for the Machine Translation task. It is hoped that in the future, there will be more resources in the realm of Cantonese Natural Language Processing, which could lead to a big improvement in different tasks related to Cantonese, including Automatic Speech Recognition, Machine Translation tasks, and others.
    \item \textbf{{Computational Resources}} Another major limitation that leads to only fine-tuning in the methodology is the lack of computing power. Training a Transformer to be fully capable of understanding the semantics and translating to a different language requires a substantial amount of computing power and data, which makes it impossible to take an existing model and conduct further pre-training.
\end{itemize}

\subsection{Future Work}
Several areas can be further investigated to enhance the system's translation quality and potentially increase the project's impact.
\begin{itemize}
    \item \textbf{Enlarging Datasets} More datasets are found during the latter stages of the project. Due to the long training time needed for 1 full iteration of the model, there is not enough time to train the best model with a more data setup. It is hoped that in the future, a new model could be trained with the newly obtained dataset at the end of the full iteration, potentially reaching a new state-of-the-art performance compared to commercial translators.
    \item \textbf{{Data Cleaning}} Another important aspect would be data cleaning. we strongly believe that with a better-quality monolingual dataset in Cantonese, there could be a significant improvement in building a better-performing model. Future work could be done where only high-quality sentences are extracted and used in the monolingual corpus rather than purely long sentences.
    \item \textbf{Model Extensions}  The investigation so far involves only fine-tuning of models, with more data available and different types of data which are not used in the current work, 2 approaches can be carried out to improve the system's performance. a), as demonstrated in other papers, a potential way of building a better system could be connecting 2 BART models (Cantonese, English) with an additional layer in between, this approach would need huge computing power since additional training is needed to fine-tune BART as a Cantonese encoder rather than a Chinese encoder. b), during the pre-processing phase, we noticed that there could be a method to convert Cantonese characters to the romanisation of Cantonese (JyutPing), where studies in Chinese Machine Translation  \citep{Du2017Pinyin} have shown that with the romanisation of Chinese(PinYin), the performance of model could be improved.
    \item \textbf{User Interface Refinement} The existing user interface is still very raw and probably incapable of scaling if third parties wish to use the application for purposes other than local usage. The web application can be further developed to be more scalable for real-life deployment purposes.
\end{itemize}

\begin{acks}
We thank {Bernice Ko, Jasmine Chan, Anson Hong} for their contributions to the human evaluation tasks.
We acknowledge the usage of open-source software NLLB, mBART, OpusMT; and the corpus from words.hk, Wenlin corpus, WMT2012, and LIHKG.
LH and GN are grateful for the grant “Integrating hospital outpatient letters into the healthcare data space” (EP/V047949/1; funder: UKRI/EPSRC).

\end{acks}

\bibliographystyle{ACM-Reference-Format}
\bibliography{sample-base}

\appendix

\section{\textsc{CantonMT} Open-sourced Toolkit}
\label{app:toolkit}
\subsection{Web Application}
\label{method:web}
For potential users to evaluate between different models, a user interface was developed to allow users to choose between different trained models and different languages.

\begin{figure}[h!]\centering
\includegraphics[width=1\linewidth]{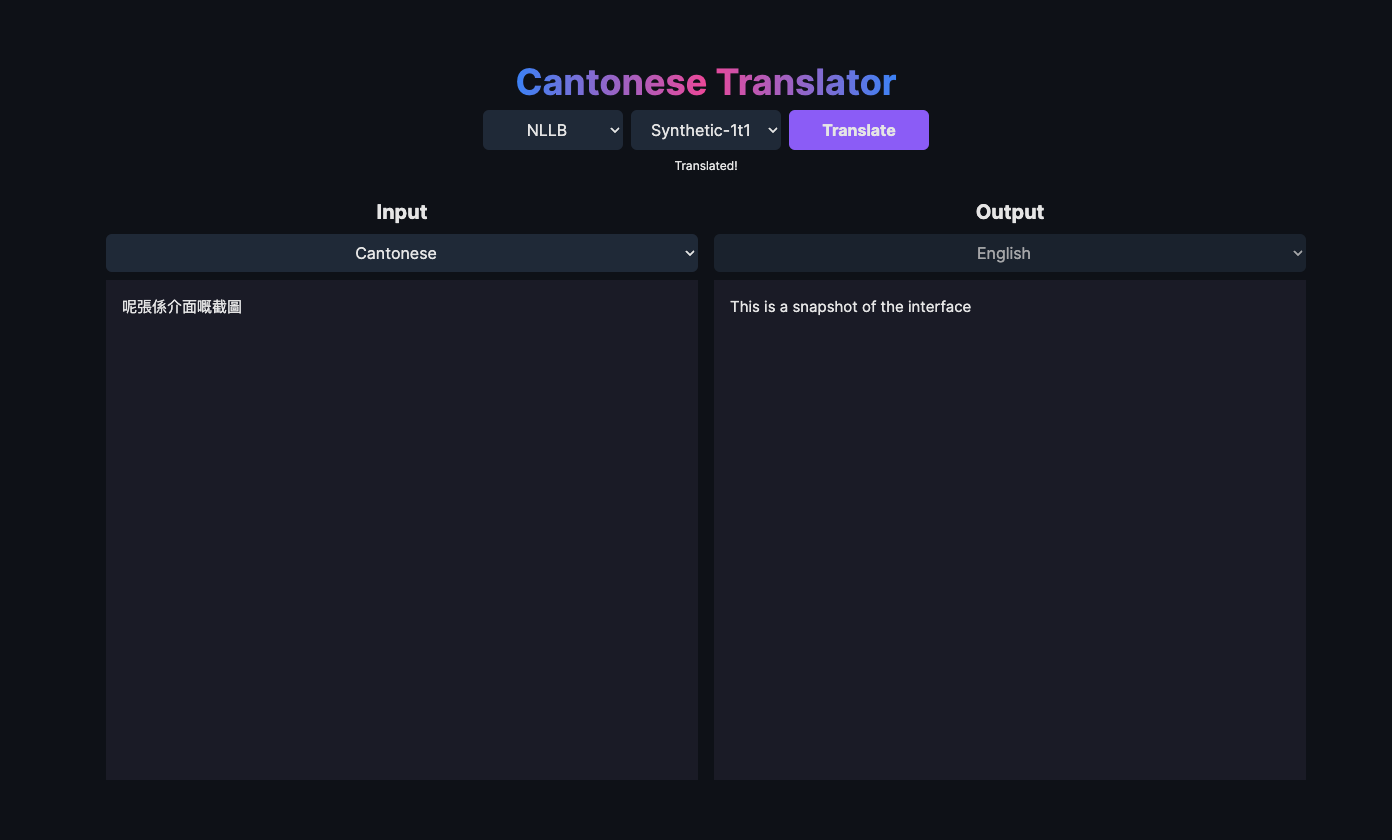}\\    
\caption{Screenshots taken from Web Application} \label{fig:UI} 
\end{figure}

The web application contains two main parts, Interface and Server, and their interactions are described in the diagram and are detailed in the following subsections. Screenshots of the web application can also be seen in Figure \ref{fig:UI}.\\

\label{appendix:UI}

\begin{figure*}[ht]
\centering
\includegraphics[width=0.99\textwidth]{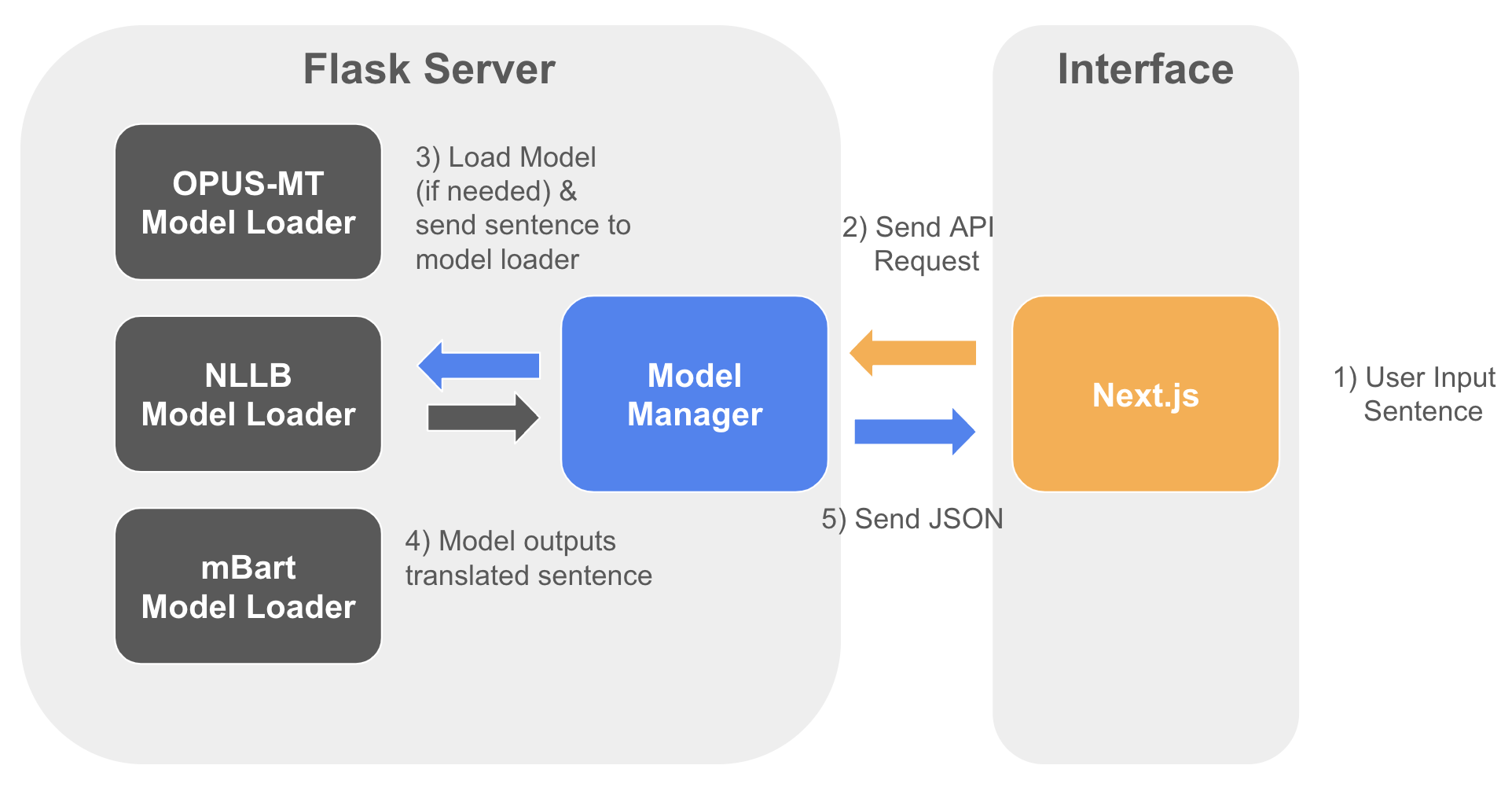}
    \caption{Diagram for CantoMT User Interface Data Flow}
    \label{fig:UI_Flow}
\end{figure*}

Figure \ref{fig:UI_Flow} outlines the general data flow for the User Interface to aid readers in understanding the structure of the server and User Interface. 

\subsubsection{User Interface}
To test out the User Interface and different models for translation, users can choose different model types and languages, which dynamically capture the available model from the server and allow users to select different training methods for the model. One can then type the sentence in the input box and click the translate button for the translation output from the model.\\

The application layout is entirely modular in case different model types or languages are added to the system, which could potentially be used as a base framework for different translation systems and add more languages to the input and output if one wishes to expand the implementations.\\

This web application has taken a template  \citep{ai-code-translator} for an AI Code translator and modified it to fit the need, which is developed in TypeScript with the Next.js framework. The reason for choosing this framework is that it provides a very modern and minimalistic approach to web development.

\subsubsection{Server}

Users can easily run the server on their local machines, and it is well-documented in the Readme file in the code submission file. The server has two main functionalities, the first will output the list of model paths given the model type and source languages. The second one provides the translation, where one could provide the details of the model and the sentence in the specified language, and the server would respond with the translated sentence using the model output.\\

Due to a lack of memory space during the implementation, the server crashed multiple times on our local machine. To account for the potential crash, a model manager was produced, which implements an LRU cache for the different model loaders, where the least recently used model will be deleted from memory if it exceeds the limit of the number of models.\\

It can also be said that the code written is very modular, which could mean that in future work, which might be a completely new model type, one can add the class for the specific model type and expand on the base model loader with minimal effort.\\

The server is built entirely on the Python Flask library. The Flask framework is chosen because the models can be run on the Python Transformers library, providing a seamless implementation without much additional effort.

\section{Evaluation Details}
\label{app:evals}

\subsection{Examples Output}

\begin{CJK*}{UTF8}{bkai}
\begin{table*}[!htb]
    \centering
    \begin{tabular}{p{0.45\textwidth}|p{0.45\textwidth}}
    \toprule
         Source Sentence & Model Translation \\\hline
         呢位小朋友智力發展遲緩。 & The intellectual development of this child is slow.\\\hline
         我漏咗個銀包喺小巴度。 & I left my wallet in the minibus.\\\hline
         嗰次交通意外令到佢變咗殘廢。 & That traffic accident left him crippled.\\\hline
         佢好爽快咁應承咗我。 & He promised me so quickly.\\\hline
         我買咗個大蛋糕慶賀亞爺七十大壽。 & I bought a big cake to celebrate my grandfather's 70th birthday.\\\hline
         呢單嘢你一定要金睛火眼幫我睇住。 & You must keep an eye on this matter for me.\\\hline
         我識咗佢十幾年喇，我地係生意上嘅好搭檔嚟。 & I've known him for over ten years. We are good business partners.\\
    \bottomrule
    \end{tabular}
    \caption{Example Outputs from Best Model}
    \label{tab:example-output}
\end{table*}
\end{CJK*}

\subsection{Human Annotation Guidelines}
\label{appendix:hopes-guide}
\begin{figure}[ht!]
 \centering
  \includegraphics[width=1.3\linewidth, angle=90]{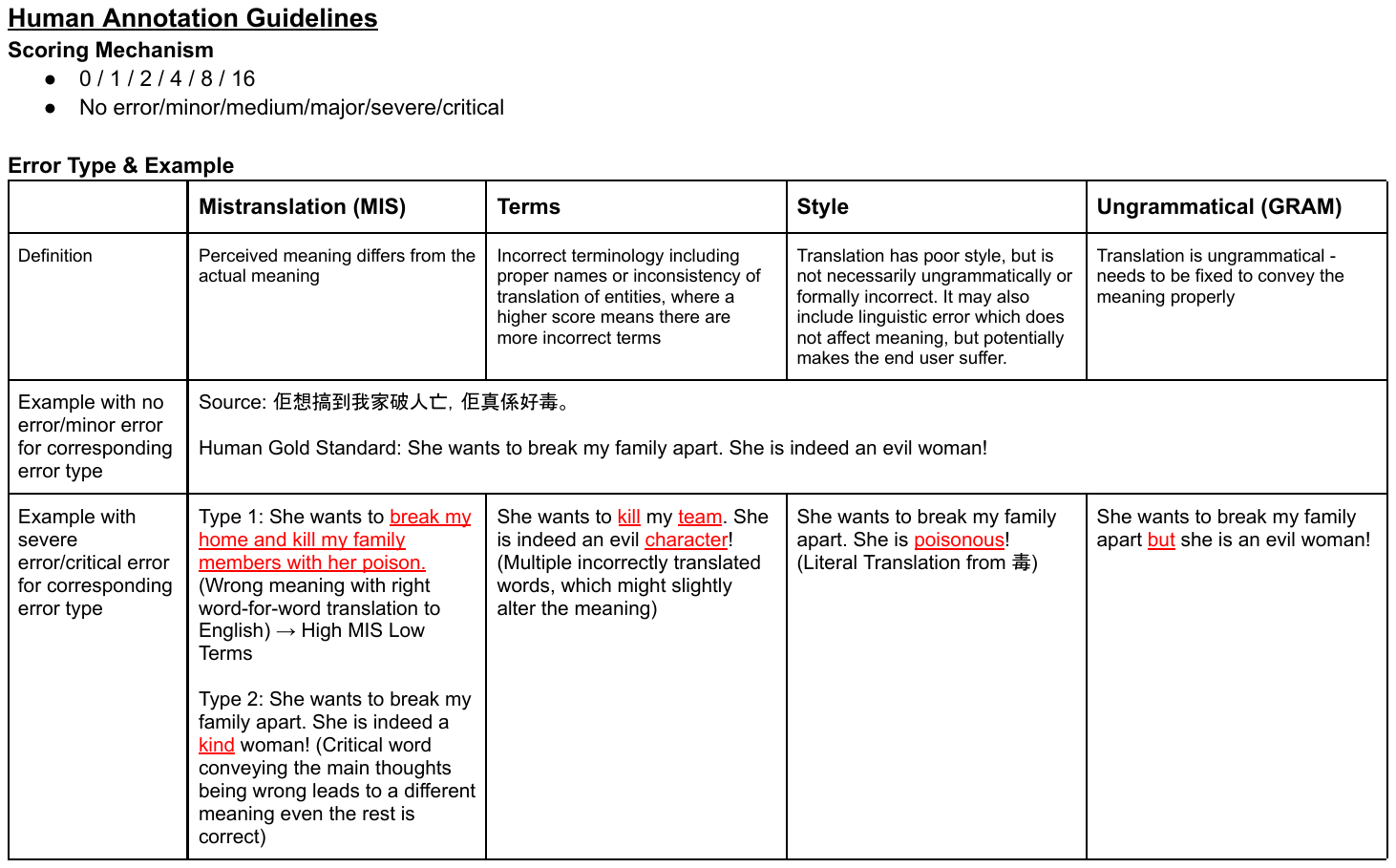}
\end{figure}
\subsection{Automatic Evaluation Results}
\label{appendix:auto-eval}
\begin{table*}[!htb]
    \centering
    \begin{tabular}{l|cccc}
        \toprule
        Model Name & SacreBLEU & hLEPOR & BERTscore & COMET \\ \hline
        nllb-forward-bl & 16.5117 & 0.5651 & 0.9248 & 0.7376 \\
        nllb-forward-syn-h:h & 15.7751 & 0.5616 & 0.9235 & 0.7342 \\ 
        nllb-forward-syn-1:1 & \textbf{16.5901} & 0.5686 & \textbf{0.925} & \textbf{0.7409} \\ 
        nllb-forward-syn-1:1-10E & 16.5203 & \textbf{0.5689} & 0.9247 & 0.738 \\ 
        nllb-forward-syn-1:3 & 15.9175 & 0.5626 & 0.924 & 0.7376 \\ 
        nllb-forward-syn-1:5 & 15.8074 & 0.562 & 0.9237 & 0.7386 \\ \hline
        nllb-forward-syn-1:1-mbart & \textbf{16.8077} & \textbf{0.571} & \textbf{0.9256} & \textbf{0.7425} \\
        nllb-forward-syn-1:3-mbart & 15.8621 & 0.5617 & 0.9246 & 0.7384 \\
        nllb-forward-syn-1:1-opus & 16.5537 & 0.5704 & 0.9254 & 0.7416 \\
        nllb-forward-syn-1:3-opus & 15.9348 & 0.5651 & 0.9242 & 0.7374 \\ \hline
        mbart-forward-bl & 15.7513 & 0.5623 & 0.9227 & 0.7314 \\
        mbart-forward-syn-1:1-nllb & \textbf{16.0358} & \textbf{0.5681} & \textbf{0.9241} & \textbf{0.738} \\
        mbart-forward-syn-1:3-nllb & 15.326 & 0.5584 & 0.9225 & 0.7319 \\ \hline
        opus-forward-bl-10E & \textbf{15.0602} & \textbf{0.5581} & \textbf{0.9219} & \textbf{0.7193} \\
        opus-forward-syn-1:1-10E-nllb & 13.0623 & 0.5409 & 0.9164 & 0.6897 \\
        opus-forward-syn-1:3-10E-nllb & 13.3666 & 0.5442 & 0.9167 & 0.6957 \\ \hline
        baidu & 16.5669 & 0.5654 & 0.9243 & 0.7401 \\ 
        bing & 17.1098 & 0.5735 & 0.9258 & 0.7474 \\ 
        gpt4-ft(CantoneseCompanion) & \textbf{19.1622} & \textbf{0.5917} & \textbf{0.936} & \textbf{0.805} \\ \hline\hline 
        nllb-forward-bl-plus-wenlin14.5k & {\textit{16.6662}} & \underline{\textit{0.5828}} & \underline{\textit{0.926}} & \underline{0.7496} \\
        mbart-forward-bl-plus-wenlin14.5k & 15.2404 & 0.5734 & 0.9238 & 0.7411 \\
        opus-forward-bl-plus-wenlin14.5k & 13.0172 & 0.5473 & 0.9157 & 0.6882 \\ \hline\hline 
        nllb-200-deploy-no-finetune & 11.1827 & 0.4925 & 0.9129 & 0.6863 \\
        opus-deploy-no-finetune & 10.4035 & 0.4773 & 0.9082 & 0.6584 \\
        mbart-deploy-no-finetune & 8.3157 & 0.4387 & 0.9005 & 0.6273 \\ \hline\hline 
        nllb-forward-all3corpus & \underline{\textit{16.9986}} & \underline{\textit{0.583}} & \underline{\textit{0.927}} & \underline{\textit{0.7549}} \\
        nllb-forward-all3corpus-10E & 16.1749 & 0.5728 & 0.9254 & 0.7508 \\
        mbart-forward-all3corpus & 16.3204 & 0.5766 & 0.9253 & 0.7482 \\
        opus-forward-all3corpus-10E & 14.4699 & 0.5621 & 0.9191 & 0.7074 \\
        \bottomrule
    \end{tabular}
    \caption{Automatic Evaluation Scores from Different Models of \textsc{ CantonMT}. 
    }
    \label{tab:Evaluation-metrics-scores}
\end{table*}

Table \ref{tab:Evaluation-metrics-scores} shows the 
Automatic Evaluation Scores from Different Models in\textsc{ CantonMT} in one view, where
bl: bilingual real data; syn: synthetic data; h:h - half and half; 1:1/3/5 - 100\% real + 100/300/500\% synthetic; 10E: 10 epochs (default: 3); top-down second slot: model switch: model type using NLLB but synthetic data from other models (mBART and OpusMT); top-down third slot: including model switch for mBART fine-tuning using synthetic data generated from NLLB; similarly top-down forth slot: including model switch for OpusMT fine-tuning using synthetic data from NLLB. Bottom slot of Cluster 1: Bing/Baidu Translator and GPT4-finetuned Cantonese Companion; \textbf{bold} case is the best score of the same slot among the same model categories.
    Cluster 2: bilingual fine-tuned models using 38K words.hk data plus 14.5k Wenlin data; \textit{italic} indicates the number outperforms the same model fine-tuned with less data 38K.
    Cluster 3: Deployed Model without fine-tuning.
    Cluster 4: Finetuned with the previous 2 corpora and an additional 10K data from OPUS Corpora we managed to find in the end - it shows the evaluation improvement continues.

\end{document}